 \documentclass[final,1p,times]{elsarticle}

\usepackage{amssymb}
\usepackage[latin1]{inputenc}
\usepackage{multirow}
\usepackage{amsmath,amsthm}
\usepackage{subfigure}
\usepackage{amsfonts}

\journal{Fuzzy Sets and Systems}

\graphicspath{}

\begin{document}

\begin{frontmatter}


\title{Characterizing Quantifier Fuzzification Mechanisms: a behavioral guide for practical applications}

\author{F. Diaz-Hermida\corref{*}}
\ead{felix.diaz@usc.es}
\cortext[*]{Corresponding author. Tel.: +0034 881816410}

\author{M. Pereira-Fariña}
\ead{martin.pereira@usc.es}

\author{Juan C.~Vidal}
\ead{juan.vidal@usc.es}

\author{A. Ramos-Soto}
\ead{alejandro.ramos@usc.es}

\address{Centro de Investigaci\'on en Tecnolox\'ias da Informaci\'on (CITIUS), University of Santiago de Compostela, Campus Vida, E-15782, Santiago de Compostela, Spain}

\begin{abstract}
Important advances have been made in the fuzzy quantification field. Nevertheless, some problems remain when we face the decision of selecting the most convenient model for a specific application. In the literature, several desirable adequacy properties have been proposed, but theoretical limits impede quantification models from simultaneously fulfilling every adequacy property that has been defined. Besides, the complexity of model definitions and adequacy properties makes very difficult for real users to understand the particularities of the different models that have been presented. In this work we will present several criteria conceived to help in the process of selecting the most adequate Quantifier Fuzzification Mechanisms for specific practical applications. In addition, some of the best known well-behaved models will be compared against this list of criteria. Based on this analysis, some guidance to choose fuzzy quantification models for practical applications will be provided.
\end{abstract}


\end{frontmatter}

\providecommand{\U}[1]{\protect\rule{.1in}{.1in}}
\newtheorem{theorem}{Theorem}
\newtheorem{defn}[theorem]{Definition}
\newtheorem{exmp}[theorem]{Example}
\newtheorem{proposition}[theorem]{Proposition}

\section{Introduction}

The evaluation of fuzzy quantified expressions is a topic that has been widely
dealt with in literature
\cite{Barro02,Delgado00,Delgado14,Sanchez16,DiazHermida00,DiazHermida02-FuzzySets,DiazHermida04IPMU,DiazHermida10Arxiv,Dubois85,Glockner03-Generalized,Glockner06Libro,Liu98,Ming2006,Ralescu95,Yager83,Yager88,Zadeh83}%
. The range of applications of fuzzy quantification includes fuzzy control
\cite{Yager84}, temporal reasoning in robotics \cite{Mucientes03}, fuzzy
databases \cite{Bosc95Sqlf}, information retrieval
\cite{Bordogna00,losada-etal04,DiazHermida04-IEEE}, data fusion
\cite{Yager88,Glockner98-Fusion} and more recently data-to-text applications
\cite{RamosSoto15,RamosSoto16}.

Moreover, the definition of adequate models to evaluate quantified expressions
is fundamental to perform `computing with words', topic that was suggested by
Zadeh \cite{Zadeh96} to express the ability of programming systems in a
linguistic way.

In general, most approaches to fuzzy quantification use the concept of
\textit{fuzzy linguistic quantifier} to represent absolute or proportional
fuzzy quantities. Zadeh \cite{Zadeh83} \textit{defined quantifiers of the
first type} as quantifiers used for representing absolute quantities (by using
fuzzy numbers on $\mathbb{N}$) , and \textit{quantifiers of the second type
}as quantifiers used for representing relative quantities (defined by using
fuzzy numbers on $\left[  0,1\right]  $).

For analyzing the behavior of fuzzy quantification models different properties
of convenient or necessary fulfillment have been defined
\cite{Delgado00,DiazHermida06Tesis,DiazHermida10Arxiv,Glockner06Libro}.
However, most of the approaches fail to exhibit a plausible behavior as has
been proved through the different reviews that have been published
\cite{Barro02,Delgado00,Delgado14,Glockner06Libro,DiazHermida06Tesis} and only
a few
\cite{Delgado00,DiazHermida02-FuzzySets,Glockner03-Generalized,Glockner06Libro}
seem to exhibit an adequate behavior in the general case.

In this work, we will follow Gl\"{o}ckner's approximation to fuzzy
quantification \cite{Glockner06Libro}. In his approach, the author generalizes
the concept of \textit{generalized classic quantifier} \cite{Barwise81}
(second order predicates or set relationships) to the fuzzy case; that is, a
\textit{fuzzy quantifier }is a fuzzy relationship between fuzzy sets. And then
he rewrites the fuzzy quantification problem as the problem of looking for a 
mechanism to transform \textit{semi-fuzzy quantifiers} (quantifiers in a
middle point between generalized classic quantifiers and fuzzy quantifiers,
used to specify the meaning of quantified expressions) into fuzzy quantifiers.
The author calls these transformation mechanisms \textit{Quantifier
Fuzzification Mechanism} (\textit{QFMs}). Being based in the linguistic
\textit{Theory of Generalized Quantifiers (TGQ)} \cite{Barwise81}, this
approach is capable of handling most of the quantification phenomena of
natural language. In addition, including quantification into a common
theoretical framework following TGQ, it also allows the translation of most of
the analysis that has been made from a linguistic perspective to the fuzzy
case, and facilitates the definition and the test of adequacy properties.

Gl\"{o}ckner has also defined a rigorous axiomatic framework to ensure the
good behavior of QFMs. Models fulfilling this framework are called
\textit{Determiner fuzzification schemes (DFSs) }and they comply with a broad
set of properties that guarantee a good behavior from a linguistic and fuzzy
point of view. See the recent \cite{Sanchez16} or \cite{Glockner06Libro} for a
comparison between Zadeh's and Gl\"{o}ckner's approaches.

The DFS framework has supposed a notable advance and several well behaved QFMs
have been identified \cite{Glockner06Libro}, \cite{DiazHermida06Tesis}.
However, important problems still remain when we must face the decision of
selecting an specific QFM for a practical application. First, it has been
proved that no model can fulfill every desirable adequacy property that has
been proposed \cite{Glockner06Libro}, and as a consequence, a `perfect model'
cannot exist. Besides, the complexity of the definition of the models and
adequacy properties makes really difficult for a user to decide which one is
the most convenient for a certain application. In addition, as we will show 
along the exposition, there are some criteria that have not been previously
taken into account for analyzing the plausible models and, even for the cases
in which some of these criteria had been previously considered, a complete
comparison among the behavior of at least the best-behaved models has not
been done.

In this work we will focus on, to the best of our knowledge, the best-behaved
QFMs: models $\mathcal{F}^{MD}$, $\mathcal{F}^{I}$, $\mathcal{F}^{A}$
\cite{DiazHermida06Tesis} and models $\mathcal{M}$, $\mathcal{M}_{CX}$ and
$\mathcal{F}_{owa}$ \cite{Glockner06Libro} with the objective of establishing
a set of criteria that facilitates the understanding of the behavioral
differences among them and helping with the process of selecting the more
convenient model for applications. All the selected models, being QFMs,
present a more general definition than models following Zadeh's framework
\cite{Zadeh83}. Furthermore, some of them generalize other known approaches,
as the ones based on the Sugeno or Choquet integrals. Thus, selected models
comprise a really good representation of the `state of the art' of fuzzy
quantification. We refer the reader to the exhaustive and recent revision in
\cite{Delgado14} for a thoroughly comparative analysis of fuzzy quantification
proposals. Previous state of the art revisions about the fuzzy quantification
field can be found in
\cite{Barro02,Delgado00,Glockner06Libro,DiazHermida06Tesis}.

Before continuing, we would remark that only the models $\mathcal{F}^{A}$,
$\mathcal{M}$, $\mathcal{M}_{CX}$ and $\mathcal{F}_{owa}$ fulfill the strict
DFS framework, being alpha-cut based models like $\mathcal{F}^{MD}$ and
$\mathcal{F}^{I}$, previously considered as non plausible from the point of
view of the DFS framework \cite[section 7.2]{Glockner06Libro}. In order to
understand the differences between these models and DFSs, we will first
compare the selected models against the main properties considered into the
QFM framework. Once the main differences derived from the properties described
in \cite{Glockner06Libro} have been presented, we will introduce the new set of
criteria that will allow us to improve the comparison between the different
models and to prove that, for some problems, alpha-cut based models
$\mathcal{F}^{MD}$ and $\mathcal{F}^{I}$ can be superior to known DFSs.

Moreover, as we will argue when we analyze the different models against the
set of criteria introduced in this paper, a `clear winner' cannot be
identified, being the general situation that some models are more appropriate
for some applications than others.

The paper is organized as follows. Section 2 will summarize Gl\"{o}ckner's
approach to fuzzy quantification, based on quantifier fuzzification
mechanisms.\ In section 3, we will present the definition of the models
$\mathcal{F}^{MD}$, $\mathcal{F}^{I}$, $\mathcal{F}^{A}$, $\mathcal{M}$,
$\mathcal{M}_{CX}$ and $\mathcal{F}_{owa}$. Section 4 will present the main
properties considered in the QFM framework \cite{Glockner06Libro} and a brief
comparison of the models $\mathcal{F}^{MD}$, $\mathcal{F}^{I}$, $\mathcal{F}%
^{A}$, $\mathcal{M}$, $\mathcal{M}_{CX}$ and $\mathcal{F}_{owa}$, with the
objective of clearly identifying the behavioral differences of these models
with respect to these properties. Section 5 will be devoted to establish the
set of criteria that will allow us to improve the comparison of the considered
models, and to analyze the different models against this new set of criteria.
Section 6 summarizes the results and establishes some criteria to guide in the
model selection for applications. The paper is closed with some conclusions.

\section{The fuzzy quantification framework}

To overcome Zadeh's framework to fuzzy quantification Gl\"{o}ckner,
\cite{Glockner06Libro} rewrote the problem of fuzzy quantification as the
problem of looking for adequate ways to convert specification means
(semi-fuzzy quantifiers) into operational means (fuzzy quantifiers).

Fuzzy quantifiers are just a fuzzy generalization of crisp or classic
quantifiers. Before giving the definition of fuzzy quantifiers, we will show
the definition of classic quantifiers according to TGQ.

\begin{defn}
A two valued (generalized) quantifier on a base set $E\neq\varnothing$ is a
mapping $Q:\mathcal{P}\left(  E\right)  ^{n}\longrightarrow\mathbf{2}$, where
$n\in\mathbb{N}$ is the arity (number of arguments) of $Q$, $\mathbf{2}%
=\left\{  0,1\right\}  $ denotes the set of crisp truth values, and
$\mathcal{P}\left(  E\right)  $ is the powerset of $E$.
\end{defn}

Examples of some definitions of classic quantifiers are:%
\begin{align*}
\mathbf{all}\left(  Y_{1},Y_{2}\right)   &  =Y_{1}\subseteq Y_{2}\\
\mathbf{at\_least}80\%\left(  Y_{1},Y_{2}\right)   &  =\left\{
\begin{array}
[c]{cc}%
\frac{\left\vert Y_{1}\cap Y_{2}\right\vert }{\left\vert Y_{1}\right\vert
}\geq0.80 & Y_{1}\neq\varnothing\\
1 & Y_{1}=\varnothing
\end{array}
\right.
\end{align*}

In a fuzzy quantifier, arguments and results can be fuzzy. A fuzzy quantifier
assigns a gradual result to each choice of $X_{1},\ldots,X_{n}\in
\widetilde{\mathcal{P}}\left(  E\right)  $, where by $\widetilde{\mathcal{P}%
}\left(  E\right)  $ we denote the fuzzy powerset of $E$.

\begin{defn}
\cite[definition 2.6]{Glockner06Libro} An n-ary fuzzy quantifier
$\widetilde{Q}$ on a base set $E\neq\varnothing$ is a mapping $\widetilde
{Q}:\widetilde{\mathcal{P}}\left(  E\right)  ^{n}\longrightarrow
\mathbf{I=}\left[  0,1\right]  $.
\end{defn}

For example, the fuzzy quantifier $\widetilde{\mathbf{all}}:\widetilde
{\mathcal{P}}\left(  E\right)  ^{2}\longrightarrow\mathbf{I}$ could be defined
as:
\[
\widetilde{\mathbf{all}}\left(  X_{1},X_{2}\right)  =\inf\left\{  \max\left(
1-\mu_{X_{1}}\left(  e\right)  ,\mu_{X_{2}}\left(  e\right)  \right)  :e\in
E\right\}
\]
where by $\mu_{X}\left(  e\right)  $ we denote the membership function of
$X\in\widetilde{\mathcal{P}}\left(  E\right)  $.

Although a certain consensus may be achieved to accept previous expression as
a suitable definition for $\widetilde{\mathbf{all}}$ this is not the unique
possible one. The problem of establishing consistent fuzzy definitions for
quantifiers (e.g., \textit{`at least eighty percent'}) is faced in
\cite{Glockner06Libro} by introducing the concept of semi-fuzzy quantifiers. A
semi-fuzzy quantifier represents a medium point between classic quantifiers
and fuzzy quantifiers. Semi-fuzzy quantifiers are similar but far more general
than Zadeh's linguistic quantifiers \cite{Zadeh83}. A semi-fuzzy quantifier
only accepts crisp arguments, as classic quantifiers, but let the result range
over the truth grade scale $\mathbf{I}$, as for fuzzy quantifiers.

\begin{defn}
\cite[definition 2.8]{Glockner06Libro} An n-ary semi-fuzzy quantifier $Q$ on a
base set $E\neq\varnothing$ is a mapping $Q:\mathcal{P}\left(  E\right)
^{n}\longrightarrow\mathbf{I}$.
\end{defn}

$Q$ assigns a gradual result to each pair of crisp sets $\left(  Y_{1}%
,\ldots,Y_{n}\right)  $. Examples of semi-fuzzy quantifiers are:
\begin{align}
\mathbf{about\_5}\left(  Y_{1},Y_{2}\right)   &  =T_{2,4,6,8}\left(
\left\vert Y_{1}\cap Y_{2}\right\vert \right) \label{about_or_more_80}\\
\mathbf{at\_least}\_\mathbf{about}80\%\left(  Y_{1},Y_{2}\right)   &
=\left\{
\begin{array}
[c]{cc}%
S_{0.5,0.8}\left(  \frac{\left\vert Y_{1}\cap Y_{2}\right\vert }{\left\vert
Y_{1}\right\vert }\right)  & X_{1}\neq\varnothing\\
1 & X_{1}=\varnothing
\end{array}
\right. \nonumber
\end{align}
where $T_{2,4,6,8}\left(  x\right)  $ and $S_{0.5,0.8}\left(  x\right)  $
represent the common trapezoidal and $S$ fuzzy numbers\footnote{Functions
$T_{a,b,c,d}$ and $S_{\alpha,\gamma}$ are defined as
\[
T_{a,b,c,d}\left(  x\right)  =\left\{
\begin{array}
[c]{cc}%
0 & x\leq a\\
\frac{x-a}{b-a} & a<x\leq b\\
1 & b<x\leq c\\
1-\frac{x-c}{d-c} & c<x\leq d\\
0 & d<x
\end{array}
\right.  \quad,S_{\alpha,\gamma}\left(  x\right)  =\left\{
\begin{tabular}
[c]{ll}%
$0$ & $x<\alpha$\\
$2\left(  \frac{\left(  x-\alpha\right)  }{\left(  \gamma-\alpha\right)
}\right)  ^{2}$ & $\alpha<x\leq\frac{\alpha+\gamma}{2}$\\
$1-2\left(  \frac{\left(  x-\gamma\right)  }{\left(  \gamma-\alpha\right)
}\right)  ^{2}$ & $\frac{\alpha+\gamma}{2}<x\leq\gamma$\\
$1$ & $\gamma<x$%
\end{tabular}
\ \ \ \ \ \ \ \ \ \ \ \ \ \ \ \ \ \right.
\]
\par
{}}.

Semi-fuzzy quantifiers are much more intuitive and easier to define than fuzzy
quantifiers, but they do not solve the problem of evaluating fuzzy quantified
sentences. In fact, additional mechanisms are needed to transform semi-fuzzy
quantifiers into fuzzy quantifiers, i.e., mappings with domain in the universe
of semi-fuzzy quantifiers and range in the universe of fuzzy quantifiers:

\begin{defn}
\cite[definition 2.10]{Glockner06Libro}A quantifier fuzzification mechanism
(QFM) $\mathcal{F}$ assigns to each semi-fuzzy quantifier $Q:\mathcal{P}%
\left(  E\right)  ^{n}\rightarrow\mathbf{I}$ a corresponding fuzzy quantifier
$\mathcal{F}\left(  Q\right)  :\widetilde{\mathcal{P}}\left(  E\right)
^{n}\rightarrow\mathbf{I}$ of the same arity $n\in\mathbb{N}$ and on the same
base set $E$.
\end{defn}

\section{Some paradigmatic QFMs\label{Apendice}}

\subsection{Standard DFSs based on trivalued
cuts\label{SubAnalisisPropuestasTrivaluadas}}

In this section we will present the three main Gl\"{o}ckner's approaches
\cite{Glockner06Libro}. All the models that have been proposed by Gl\"{o}ckner
are standard \textit{Determiner Fuzzification Schemes}, and as a consequence
they show an excellent theoretical behavior. These models are called standard
because they induce the standard $tnorm$ $min$ and the standard $tconorm$
$max$.\footnote{In \cite[section 3.4]{Glockner06Libro} it is explained how
semi-fuzzy quantifiers can be used\ `to embed' the classical logical
functions. By means of the application of a QFM $\mathcal{F}$, we can study if
$\mathcal{F}$ transforms the classical logical functions into approppriate
fuzzy logical functions.}

Before presenting models $\mathcal{M}$ \cite[definition 7.22]{Glockner06Libro}%
, $\mathcal{M}_{CX}$ \cite[definition 7.56]{Glockner06Libro} and
$\mathcal{F}_{owa}$ \cite[definition 8.13]{Glockner06Libro} we need to
introduce some additional definitions.

\begin{defn}
\cite[definition 7.15]{Glockner06Libro}Let $E$ be a referential set,
$X\in\widetilde{\mathcal{P}}\left(  E\right)  $ a fuzzy set, and $\gamma
\in\mathbf{I}.$ $X_{\gamma}^{\min},X_{\gamma}^{\max}\in\mathcal{P}\left(
E\right)  $ are defined as:%
\begin{align*}
X_{\gamma}^{\min}  &  =\left\{
\begin{tabular}
[c]{lll}%
$X_{>\frac{1}{2}}$ & $:$ & $\gamma=0$\\
$X_{\geq\frac{1}{2}+\frac{1}{2}\gamma}$ & $:$ & $\gamma>0$%
\end{tabular}
\right. \\
X_{\gamma}^{\max}  &  =\left\{
\begin{tabular}
[c]{lll}%
$X_{\geq\frac{1}{2}}$ & $:$ & $\gamma=0$\\
$X_{>\frac{1}{2}-\frac{1}{2}\gamma}$ & $:$ & $\gamma>0$%
\end{tabular}
\right.
\end{align*}
where $X_{\geq\alpha}=\left\{  e\in E:\mu_{X}\left(  e\right)  \geq
\alpha\right\}  $ is the alpha-cut of level $\alpha$ of $X$ and $X_{>\alpha
}=\left\{  e\in E:\mu_{X}\left(  e\right)  >\alpha\right\}  $ is the strict
alpha-cut of level $\alpha$.
\end{defn}

In previous expression, $X_{\gamma}^{\min}$ represents the elements that
without doubt, belong to the fuzzy set $X$ for the `cautiousness' level
$\gamma$ whilst $X_{\gamma}^{\max}$ includes also the elements whose
belongingness to the cautiousness level $\gamma$ is undefined. Elements that
are not in $X_{\gamma}^{\max}$ do not belong to the cautiousness level
$\gamma$. The cautiousness cut can be interpreted as a `trivalued set'\ in
which elements in $X_{\gamma}^{\min}$ have membership function of $1$, whilst
for elements in $X_{\gamma}^{\max}/X_{\gamma}^{\min}$ belongingness is
undefined (membership degree of $\frac{1}{2}$). Membership function for
elements that are not in $X_{\gamma}^{\max}$ is $0$.

For the definition of $\mathcal{M},$ $\mathcal{M}_{CX}$ and $\mathcal{F}%
_{owa}$ we also need the fuzzy median operator:

\begin{defn}
Fuzzy median $med_{\frac{1}{2}}:\mathbf{I\times I\longrightarrow I}$ is
defined as:%
\[
med_{\frac{1}{2}}\left(  u_{1},u_{2}\right)  =\left\{
\begin{tabular}
[c]{lll}%
$\min\left(  u_{1},u_{2}\right)  $ & $:$ & $\min\left(  u_{1},u_{2}\right)
>\frac{1}{2}$\\
$\max\left(  u_{1},u_{2}\right)  $ & $:$ & $\max\left(  u_{1},u_{2}\right)
<\frac{1}{2}$\\
$\frac{1}{2}$ & $:$ & otherwise
\end{tabular}
\ \ \right.
\]

\end{defn}

The following definitions extends the fuzzy median to fuzzy sets:

\begin{defn}
Operator $m_{\frac{1}{2}}:\mathcal{P}\left(  \mathbf{I}\right)  \rightarrow
\mathbf{I}$ is defined as%
\[
m_{\frac{1}{2}}X=med_{\frac{1}{2}}\left(  \inf X,\sup X\right)
\]
for all $X\in\mathcal{P}\left(  \mathbf{I}\right)  $.
\end{defn}

The set that contains all the possible images of a quantifier over the range
defined by a three valued cut of level $\gamma$ is defined as \cite[page
100]{Glockner03-Generalized}:

\begin{defn}
Let $Q:\mathcal{P}\left(  E\right)  \rightarrow\mathbf{I}$ be a semi-fuzzy
quantifier, $X_{1},\ldots,X_{n}\in\widetilde{\mathcal{P}}\left(  E\right)  $
fuzzy sets and $\gamma\in\left[  0,1\right]  $ a cautiousness level.
$S_{Q,X_{1},\ldots,X_{n}}\left(  \gamma\right)  :\left[  0,1\right]
\rightarrow\mathbf{I}$ is defined as:%
\[
S_{Q,X_{1},\ldots,X_{n}}\left(  \gamma\right)  \left(  X_{1},\ldots
,X_{n}\right)  =\left\{  Q\left(  Y_{1},\ldots,Y_{n}\right)  :\left(
X_{i}\right)  _{\gamma}^{\min}\subseteq Y_{i}\subseteq\left(  X_{i}\right)
_{\gamma}^{\max}\right\}
\]

\end{defn}

Supremum and infimum of $S_{Q,X_{1},\ldots,X_{n}}\left(  \gamma\right)  $ are
represented by means of the following notation:

\begin{defn}
Let $Q:\widetilde{\mathcal{P}}\left(  E\right)  \rightarrow\mathbf{I}$ be a
semi-fuzzy quantifier, $X_{1},\ldots,X_{n}\in\widetilde{\mathcal{P}}\left(
E\right)  $ fuzzy sets and $\gamma\in\left[  0,1\right]  $ a cautiousness
level. $\top_{Q,X_{1},\ldots,X_{n}}\left(  \gamma\right)  :\left[  0,1\right]
\rightarrow\mathbf{I}$ is defined as:%
\[
\top_{Q,X_{1},\ldots,X_{n}}\left(  \gamma\right)  =\sup S_{Q,X_{1}%
,\ldots,X_{n}}\left(  \gamma\right)
\]

\end{defn}

\begin{defn}
Let $Q:\widetilde{\mathcal{P}}\left(  E\right)  \rightarrow\mathbf{I}$ be a
semi-fuzzy quantifier, $X_{1},\ldots,X_{n}\in\widetilde{\mathcal{P}}\left(
E\right)  $ fuzzy sets and $\gamma\in\left[  0,1\right]  $ a cautiousness
level. $\bot_{Q,X_{1},\ldots,X_{n}}\left(  \gamma\right)  :\left[  0,1\right]
\rightarrow\mathbf{I}$ is defined as:%
\[
\bot_{Q,X_{1},\ldots,X_{n}}\left(  \gamma\right)  =\inf S_{Q,X_{1}%
,\ldots,X_{n}}\left(  \gamma\right)
\]

\end{defn}

Using previous definitions we present the three paradigmatic DFSs :

\begin{defn}
\cite[definition 7.22]{Glockner06Libro} Standard DFS $\mathcal{M}:\left(
Q:\mathcal{P}\left(  E\right)  \rightarrow\mathbf{I}\right)  \rightarrow$
$(\widetilde{Q}:$ $\widetilde{\mathcal{P}}\left(  E\right)  \rightarrow
\mathbf{I)}$ is defined as%
\[
\mathcal{M}\left(  Q\right)  \left(  X_{1},\ldots,X_{n}\right)  =\int_{0}%
^{1}med_{\frac{1}{2}}\left(  \top_{Q,X_{1},\ldots,X_{n}}\left(  \gamma\right)
,\bot_{Q,X_{1},\ldots,X_{n}}\left(  \gamma\right)  \right)  d\gamma
\]

\end{defn}

\begin{defn}
\cite[definition 7.56, theorem 7.87]{Glockner06Libro} Standard DFS
$\mathcal{M}_{CX}:\left(  Q:\mathcal{P}\left(  E\right)  \rightarrow
\mathbf{I}\right)  \rightarrow\left(  \widetilde{Q}:\widetilde{\mathcal{P}%
}\left(  E\right)  \rightarrow\mathbf{I}\right)  $ is defined as%
\[
\mathcal{M}_{CX}\left(  Q\right)  \left(  X_{1},\ldots,X_{n}\right)
=\sup\left\{  Q_{V,W}^{L}\left(  X_{1},\ldots,X_{n}\right)  :V_{1}\subseteq
W_{1},\ldots,V_{n}\subseteq W_{n}\right\}
\]
where
\begin{align*}
Q_{V,W}^{L}\left(  X_{1},\ldots,X_{n}\right)   &  =\min\left(  \Xi
_{V,W}\left(  X_{1},\ldots,X_{n}\right)  ,\inf\left\{  Q\left(  Y_{1}%
,\ldots,Y_{n}\right)  :V_{i}\subseteq Y_{i}\subseteq W_{i}\right\}  \right) \\
\Xi_{V,W}\left(  X_{1},\ldots,X_{n}\right)   &  =\min_{i=1}^{n}\min\left(
\inf\left\{  \mu_{X_{i}}\left(  e\right)  :e\in V_{i}\right\}  ,\inf\left\{
1-\mu_{X_{i}}\left(  e\right)  :e\notin W_{i}\right\}  \right)
\end{align*}

\end{defn}

\begin{defn}
\cite[definition 8.13]{Glockner06Libro} Standard DFS $\mathcal{F}%
_{owa}:\left(  Q:\mathcal{P}\left(  E\right)  \rightarrow\mathbf{I}\right)
\rightarrow\left(  \widetilde{Q}:\widetilde{\mathcal{P}}\left(  E\right)
\rightarrow\mathbf{I}\right)  $ is defined as
\[
\mathcal{F}_{owa}\left(  Q\right)  \left(  X_{1},\ldots,X_{n}\right)
=\frac{1}{2}\int_{0}^{1}\top_{Q,X_{1},\ldots,X_{n}}\left(  \gamma\right)
d\gamma+\frac{1}{2}\int_{0}^{1}\bot_{Q,X_{1},\ldots,X_{n}}\left(
\gamma\right)  d\gamma
\]

\end{defn}

\subsection{Alpha-cut based QFMs\textbf{ }$\mathcal{F}^{I}$ and $\mathcal{F}%
^{MD}$}

Now, we will present the two QFMs based on alpha-cuts $\mathcal{F}^{I}$ and
$\mathcal{F}^{MD}$.

\begin{defn}
\cite[section 2.1]{DiazHermida02-FuzzySets}, \cite[chapter 3]%
{DiazHermida06Tesis}Let $Q:\mathcal{P}\left(  E\right)  ^{n}\rightarrow
\mathbf{I}$ be a semi-fuzzy quantifier over a base set $E$. The QFM
$\mathcal{F}^{MD}$ is defined as:
\[
\mathcal{F}^{MD}\left(  Q\right)  \left(  X_{1},\ldots,X_{n}\right)  =\int
_{0}^{1}Q\left(  \left(  X_{1}\right)  _{\geq\alpha},\ldots,\left(
X_{n}\right)  _{\geq\alpha}\right)  d\alpha
\]
for every $X_{1},\ldots,X_{n}\in\widetilde{\mathcal{P}}\left(  E\right)  $.
\end{defn}

When fuzzy sets $X_{1},\ldots,X_{n}\in\widetilde{\mathcal{P}}\left(  E\right)
$ are normalized and we limit ourselves to the unary and binary quantifiers
considered in the Zadeh's framework, $\mathcal{F}^{MD}$ coincides with the
quantification model $GD$ defined in \cite[page 281]{Delgado99}, \cite[section
3.3.2. and section 3.4.1.]{Sanchez99}, \cite[page 37]{Delgado00}. In this way,
$\mathcal{F}^{MD}$ generalizes the $GD$ model to the Gl\"{o}ckner's framework.

Let us now present the definition of the $\mathcal{F}^{I}$ model.

\begin{defn}
\cite{DiazHermida00}, \cite[section 2.2]{DiazHermida02-FuzzySets}%
,\cite[chapter 3]{DiazHermida06Tesis} Let $Q:\mathcal{P}\left(  E\right)
^{n}\rightarrow\mathbf{I}$ be a semi-fuzzy quantifier over a base set $E$. The
QFM $\mathcal{F}^{I}$ is defined as:
\[
\mathcal{F}^{I}\left(  Q\right)  \left(  X_{1},\ldots,X_{n}\right)  =\int
_{0}^{1}\ldots\int_{0}^{1}Q\left(  \left(  X_{1}\right)  _{\geq\alpha_{1}%
},\ldots,\left(  X_{n}\right)  _{\geq\alpha_{n}}\right)  d\alpha_{1}\ldots
d\alpha_{n}%
\]
for every $X_{1},\ldots,X_{n}\in\widetilde{\mathcal{P}}\left(  E\right)  $.
\end{defn}

\subsection{Non standard DFS $\mathcal{F}^{A}$}

The definition of the QFM $\mathcal{F}^{A}$ is based on a probabilistic
interpretation of fuzzy sets in which we interpret membership degrees as
probabilities \cite{DiazHermida06Tesis},\cite{DiazHermida10Arxiv}. However,
the $\mathcal{F}^{A}$ model can also be defined by means of fuzzy operators
without any reference to probability theory.

The QFM $\mathcal{F}^{A}$ fulfills the axioms of the DFS framework but it is
not a standard DFS, as the logic operators induced by the $\mathcal{F}^{A}$
model are the product \textit{$tnorm$} and the probabilistic sum
\textit{$tconorm$}.

\begin{defn}
Let $X\in\widetilde{\mathcal{P}}\left(  E\right)  $ be a fuzzy set, $E$
finite. The probability of the crisp set $Y\in\mathcal{P}\left(  E\right)  $
of being a representative of the fuzzy set $X\in\widetilde{\mathcal{P}}\left(
E\right)  $ is defined as%
\[
m_{X}\left(  Y\right)  =%
{\displaystyle\prod\limits_{e\in Y}}
\mu_{X}\left(  e\right)
{\displaystyle\prod\limits_{e\in E\backslash Y}}
\left(  1-\mu_{X}\left(  e\right)  \right)
\]

\end{defn}

As we have stated above, it is possible to make a similar definition without
making any reference to probability theory. If we consider the product $tnorm$
($\wedge\left(  x_{1},x_{2}\right)  =x_{1}\cdot x_{2}$) and the Lukasiewicz
implication then $m_{X}\left(  Y\right)  $ is the \textit{equipotence}%
\ between $Y$ and $X$ \cite{Bandler80}:%
\[
Eq\left(  Y,X\right)  =\wedge_{e\in E}\left(  \mu_{X}\left(  e\right)
\rightarrow\mu_{Y}\left(  e\right)  \right)  \wedge\left(  \mu_{Y}\left(
e\right)  \rightarrow\mu_{X}\left(  e\right)  \right)
\]

Using the previous definition we define the $\mathcal{F}^{A}$ DFS as:

\begin{defn}
\cite[pag. 1359]{DiazHermida04IPMU}Let $Q:\mathcal{P}\left(  E\right)
^{n}\rightarrow\mathbf{I}$ be a semi-fuzzy quantifier, $E$ finite. The DFS
$\mathcal{F}^{A}$ is defined as%
\begin{equation}
\mathcal{F}^{A}\left(  Q\right)  \left(  X_{1},\ldots,X_{n}\right)
=\sum_{Y_{1}\in\mathcal{P}\left(  E\right)  }\ldots\sum_{Y_{n}\in
\mathcal{P}\left(  E\right)  }m_{X_{1}}\left(  Y_{1}\right)  \ldots m_{X_{n}%
}\left(  Y_{n}\right)  Q\left(  Y_{1},\ldots,Y_{n}\right)
\label{ModeloVerosimilitudes}%
\end{equation}
for all $X_{1},\ldots,X_{n}\in\widetilde{\mathcal{P}}\left(  E\right)  $.
\end{defn}

The next expression is an alternative definition of the model $\mathcal{F}%
^{A}$:%
\[
\mathcal{F}^{A}\left(  X_{1},\ldots,X_{n}\right)  =%
{\displaystyle\bigvee\limits_{Y_{1}\in\mathcal{P}\left(  E\right)  }}
\ldots%
{\displaystyle\bigvee\limits_{Y_{n}\in\mathcal{P}\left(  E\right)  }}
Eq\left(  Y_{1},X_{1}\right)  \wedge\ldots\wedge Eq\left(  Y_{n},X_{n}\right)
\wedge Q\left(  Y_{1},\ldots,Y_{n}\right)
\]
where $\vee$ the Lukasiewicz $tconorm$ ($\vee\left(  x_{1},x_{2}\right)
=\min\left(  x_{1}+x_{2},1\right)  $), $\wedge$ is the product $tnorm$
($\wedge\left(  x_{1},x_{2}\right)  $ $=x_{1}\cdot x_{2}$) and $Eq\left(
Y,X\right)  $ is the equipotence between the crisp set $Y$ and the fuzzy set
$X$. In this way, definition of $\mathcal{F}^{A}$ can be done by means of
common fuzzy operators.

\section{The DFS axiomatic framework}

In this section we will present the DFS axiomatic framework
\cite{Glockner06Libro}. In the previous reference, the author has dedicated
the whole third and fourth chapters to describe the framework and the
properties that are consequence of it. For the sake of brevity, we will only
give a general overview of the framework and some intuitions about the set of
properties derived from it. We refer the reader to the previous publication
for full detail.

\begin{defn}
A QFM $\mathcal{F}$ is called a determiner fuzzification scheme (DFS) if the
following conditions are satisfied for all semi-fuzzy quantifiers
$Q:\mathcal{P}\left(  E\right)  ^{n}\rightarrow\mathbf{I}$:
\end{defn}

\begin{tabular}
[c]{|l|l|l|}\hline
Correct generalization & $\mathcal{U}\left(  \mathcal{F}\left(  Q\right)
\right)  =Q$\quad if $n\leq1$ & (Z-1)\\\hline
Projection quantifiers & $\mathcal{F}\left(  Q\right)  =\widetilde{\pi_{e}}%
$\quad if $Q=\pi_{e}$ for some $e\in E$ & (Z-2)\\\hline
Dualisation & $\mathcal{F}\left(  Q\widetilde{\square}\right)  =\mathcal{F}%
\left(  Q\right)  \widetilde{\square}$\quad$n>0$ & (Z-3)\\\hline
Internal joins & $\mathcal{F}\left(  Q\cup\right)  =\mathcal{F}\left(
Q\right)  \widetilde{\cup}$\quad$n>0$ & (Z-4)\\\hline
Preservation of monotonicity & If $Q$ is nonincreasing in the $n$-th arg,
then & (Z-5)\\\hline
& $\mathcal{F}\left(  Q\right)  $ is nonincreasing in $n$-th arg, $n>0$ &
\\\hline
Functional application & $\mathcal{F}\left(  Q\circ\underset{i=1}{\overset
{n}{\times}}\widehat{f_{i}}\right)  =\mathcal{F}\left(  Q\right)
\circ\underset{i=1}{\overset{n}{\times}}\widehat{\mathcal{F}}\left(
f_{i}\right)  $ & (Z-6)\\\hline
& where $f_{1},\ldots,f_{n}:E^{\prime}\rightarrow E,E^{\prime}\neq\varnothing$
& \\\hline
\end{tabular}

\subsection{Main properties derived from the DFS framework}

We will only make a brief exposition of the main properties derived from the
DFS framework. Full detail can be found in the aforementioned reference
\cite[chapters three and four.]{Glockner06Libro}.

\begin{itemize}
\item \textbf{Correct generalization (P1):} perhaps, the most important
property derived from the DFS framework is the \textit{correct generalization}
property. Correct generalization requires that the behavior of a fuzzy
quantifier $\mathcal{F}\left(  Q\right)  $ over crisp arguments is the
expected; that is, results obtained with a fuzzy quantifier $\mathcal{F}%
\left(  Q\right)  $ and with the corresponding semi-fuzzy quantifier $Q$ must
coincide over crisp arguments. It is included in the DFS axiomatic for
semi-fuzzy quantifiers of arities 0 and 1 (Z-1).

\item \textbf{Quantitativity (P2):} quantitative quantifiers do not depend on
any particular characteristic of the elements of the base set. In the finite
case, quantitative quantifiers can always be defined as a function of the
cardinality of the boolean combinations of the argument sets. A QFM
$\mathcal{F}$ preserves quantitativity if quantitative semi-fuzzy quantifiers
are translated into quantitative fuzzy quantifiers by $\mathcal{F}$.

\item \textbf{Projection quantifier (P3):} Axiom Z-2 guarantees that the
\textit{projection crisp quantifier} $\pi_{e}\left(  Y\right)  $ (that returns
$1$ if $e\in Y$ and $0$ in other case) is generalized to the \textit{fuzzy
projection quantifier} $\widetilde{\pi_{e}}\left(  X\right)  $ (that returns
$\mu_{X}\left(  e\right)  $).

\item \textbf{Induced propositional logic (P4):} we will say that a QFM comply
with the induced propositional logic if crisp logical functions ($\lnot\left(
x\right)  $, $\wedge\left(  x_{1},x_{2}\right)  $, $\vee\left(  x_{1}%
,x_{2}\right)  $, $\rightarrow\left(  x_{1},x_{2}\right)  $), that can be
embedded into the definition of semi-fuzzy quantifiers, are generalized to
acceptable fuzzy logical functions; that is, a negation operator, a $tnorm$, a
$tconorm$ and a fuzzy implication function.

\item \textbf{External negation (P5): }in the common case, external negation
of a semi-fuzzy quantifier is computed by the application of the standard
negation $\widetilde{\lnot}\left(  x\right)  =1-x$. A QFM fulfilling the
external negation property guarantees that $\mathcal{F}\left(  \widetilde
{\lnot}Q\right)  $ is equivalent to $\widetilde{\lnot}\mathcal{F}\left(
Q\right)  $. Thanks to the external negation property, equivalence of
expressions \textit{\textquotedblleft it is false that at least 80\% of the
hard workers are well paid\textquotedblright\ }and \textit{\textquotedblleft
less than 80\% of the hard workers are well paid\textquotedblright\ }is assured.

\item \textbf{Internal negation (P6):} the internal negation (antonym) of a
semi-fuzzy quantifier is defined as $Q\mathbf{\lnot}\left(  Y_{1},\ldots
,Y_{n}\right)  =Q\left(  Y_{1},\ldots,\mathbf{\lnot}Y_{n}\right)  $. For
example, \textit{`no'} is the antonym of \textit{`all'} because $\mathbf{all}%
\left(  Y_{1},Y_{2}\right)  \lnot=\mathbf{all}\left(  Y_{1},\lnot
Y_{2}\right)  =\mathbf{no}\left(  Y_{1},Y_{2}\right)  $. Fulfillment of the
internal negation property assures that internal negation transformation are
translated to the fuzzy case.

\item \textbf{Dualisation (P7):} the dualisation property coincides with the
Z-3 axiom of the DFS framework, being a consequence of the simultaneous
fulfillment of the external negation and internal negation properties. In
conjunction with previous properties, equivalences in the `Aristotelian
square' are maintained in the fuzzy case. As an example, equivalence of
$\mathcal{F}\left(  \mathbf{all}\right)  \left(  \mathbf{hard\_workers}%
,\mathbf{well\_paid}\right)  $ and $\mathcal{F}\left(  \mathbf{no}\right)
\left(  \mathbf{hard\_workers},\widetilde{\lnot}\mathbf{well\_paid}\right)  $
is assured, or in words, \textit{\textquotedblleft all hard workers are well
paid\textquotedblright} is equivalent to \textit{\textquotedblleft no hard
worker is not well paid\textquotedblright.}

\item \textbf{Union/intersection of arguments (P8)}: this property guarantees
the compliance with some transformations that allow to construct new
quantifiers by means of unions (and intersections) of arguments. As a
particular case, the equivalence between absolute unary and binary quantifiers
is a consequence of this axiom. As an example, the equivalence between
\textit{\textquotedblleft about 5 hard workers are well paid\textquotedblright%
} and \textit{\textquotedblleft about 5 people are hard workers and well
paid\textquotedblright\ }is assured. For QFMs fulfilling the DFS framework,
the fulfillment of this property, in combination with internal and external
negation properties, allow the preservation of the boolean argument structure
that can be expressed in natural language when none of the boolean variables
$X_{i}$ occurs more than once \ \cite[section 3.6]{Glockner06Libro}.

\item \textbf{Coherence with standard quantifiers (P9): }by standard
quantifiers we refer to the classical quantifiers $\exists,\forall$ and their
binary versions \textbf{some} and \textbf{all}. We will say that a QFM
maintains coherence with standard quantifiers if the fuzzy versions of the
classical quantifiers are the expected. For example, a QFM fulfilling this
property complies (where $\widetilde{\vee},\widetilde{\wedge},\widetilde
{\rightarrow}$ are the logical operators induced by the QFM):%
\begin{align*}
\mathcal{F}\left(  \exists\right)  \left(  X\right)   &  =\sup\left\{
\overset{m}{\underset{i=1}{\widetilde{\vee}}}\mu_{X}\left(  a_{i}\right)
:A=\left\{  a_{1},\ldots,a_{m}\right\}  \in\mathcal{P}\left(  E\right)
,a_{i}\neq a_{j}\text{ if }i\neq j\right\} \\
\mathcal{F}\left(  \mathbf{all}\right)  \left(  X_{1},X_{2}\right)   &
=\inf\left\{  \overset{m}{\underset{i=1}{\widetilde{\wedge}}}\mu_{X_{1}%
}\left(  a_{i}\right)  \widetilde{\rightarrow}\mu_{X_{2}}\left(  a_{i}\right)
:A=\left\{  a_{1},\ldots,a_{m}\right\}  \in\mathcal{P}\left(  E\right)
,a_{i}\neq a_{j}\text{ if }i\neq j\right\}
\end{align*}

\item \textbf{Monotonicity in arguments (P10): }this property assures the
translation of monotonicity in arguments relations from the semi-fuzzy to the
fuzzy case. As an example, the binary semi-fuzzy quantifier \textit{`most'} is
increasing in its second argument (e.g. \textit{\textquotedblleft most
students are poor\textquotedblright}). This property assures that the fuzzy
version of \textit{`most'} is also increasing in its second argument.

\item \textbf{Monotonicity in quantifiers (P11):} this property assures the
preservation of monotonicity relations in quantifiers. For example,
\textit{`between 4 and 6'} is more specific than \textit{`between 2 and 8'}.
Fulfilment of this property assures that in the fuzzy case, the specificity
relations between quantifiers are preserved.

\item \textbf{Crisp argument insertion (P12):} For a semi-fuzzy quantifier
$Q:\mathcal{P}\left(  E\right)  ^{n}\rightarrow\mathbf{I}$, crisp argument
insertion allow to construct a new quantifier $Q:\mathcal{P}\left(  E\right)
^{n-1}\rightarrow\mathbf{I}$ by means of the restriction of $Q$ by a crisp set
$A$; that is, the crisp argument insertion $Q\lhd A$ is defined as $Q\lhd
A\left(  Y_{1},\ldots,Y_{n-1}\right)  =Q\left(  Y_{1},\ldots,Y_{n-1},A\right)
$. A $QFM$ preserving the property of crisp argument insertion assures that
$\mathcal{F}\left(  Q\lhd A\right)  =\mathcal{F}\left(  Q\right)  \lhd A$;
that is, it is equivalent to first restrict the semi-fuzzy quantifier $Q$ by
$A$ and then applying the fuzzification scheme $\mathcal{F}$ or to first
applying the fuzzification mechanism and then restricting the corresponding
fuzzy quantifier by $A$. Crisp argument insertion allow to model
the\ `adjectival restriction' of natural language in the crisp case.
\end{itemize}

\subsection{Some relevant properties non included in the DFS framework}

In \cite[chapter six]{Glockner06Libro} some additional adequacy properties for
characterizing DFSs were described. These additional properties were not
included in the DFS framework in some cases, for being incompatible with it,
and in other cases, in order to not excessively restrict the set of
theoretical models fulfilling the framework, which was important for the
author for studying the full set of classes of standard models and their
theoretical limits. We will present now the more relevant:

\begin{itemize}
\item \textbf{Continuity in arguments (P13):} this property assures the
continuity of the models with respect to the argument sets. It is fundamental
to guarantee that small modifications in arguments do not provoke high
variations in the results of evaluating quantified expressions.

\item \textbf{Continuity in quantifiers (P14):} this property assures the
continuity of the models with respect to variations in the quantifiers.

\item \textbf{Propagation of fuzziness (P15): }this property assures that
fuzzier inputs (understood as fuzzier input sets) and fuzzier quantifiers
produce fuzzier outputs. We will discuss this property in more detail when we
introduce the set of criteria we will use to improve the characterization of
the behavior of the QFMs (see section \ref{SectionPropagationFuzziness}).

\item \textbf{Fuzzy argument insertion (P16): }this property is the fuzzy
counterpart of the crisp argument insertion. It is a very restrictive
property, that will impose great limitations into the set of models fulfilling
the DFS axiomatic framework .
\end{itemize}

\subsection{Comparison of the models against the QFM properties}

In this section we will make a brief summary of the theoretical analysis of
the QFMs $\mathcal{F}^{MD}$, $\mathcal{F}^{I}$, the non-standard
DFS\ $\mathcal{F}^{A}$ and the standard DFSs $\mathcal{M}$, $\mathcal{M}_{CX}$
and $\mathcal{F}_{owa}$ with respect to the set of previous properties. Table
\ref{TableSummaryGlockner} summarizes the fulfillment of the properties for
the different models. A detailed analysis of these models can be found in
\cite{Glockner06Libro} and \cite{DiazHermida06Tesis}.

This analysis will allow us to understand the main differences between the
models we are considering. As we will argue in the following section, although
the set of properties previously presented allow for a deep analysis of the
models, we consider that they are not enough to understand the behavioral
differences between them and to decide which ones can be more appropriate for
specific applications. The introduction of these new criteria and the analysis
of the behavior of the modes with respect to it will be the objective of the
last two sections of this paper.

Before summarizing the behavior of these models against these properties, we
would like to emphasize that our point of view is that although models
$\mathcal{F}^{MD}$ and $\mathcal{F}^{I}$ are not DFSs, they are really
competitive with respect to models fulfilling the DFS framework. The main
differences between the $\mathcal{F}^{MD}$ and $\mathcal{F}^{I}$ models when
we compare them with DFSs is that they fail to fulfill some of the linguistic
properties derived from the DFS framework. In addition, these models also fail
to fulfill some of the QFM properties in the infinite case, which do not
affect to most of the practical applications of fuzzy
quantification\footnote{We have the hypothesis that for `practical
quantifiers' (i.e., defined by means of continuous fuzzy numbers) models
$\mathcal{F}^{MD}$ and $\mathcal{F}^{I}$ fulfill the continuous in arguments
property. We also have the hypothesis that model $\mathcal{F}^{I}$ fulfill the
internal negation property for infinite domains in the same cases. The
fulfillment of these properties will guarantee the convenience of these models
for infinite domains in the practical cases.}. Finally $\mathcal{F}^{MD}$ and
$\mathcal{F}^{I}$ only fulfill the coherence with the standard quantifiers
property in the unary case, although in the specific case of the
$\mathcal{F}^{I}$ model fulfillment of the property depends on the mechanism
we will use to compute the induced operators of the model \cite[chapter
3]{DiazHermida06Tesis}.

The competitiveness of $\mathcal{F}^{MD}$ and $\mathcal{F}^{I}$ models with
respect to DFSs will become more clear when we present the analysis of the
models against the new set of criteria, which from our point of view will
prove that in some cases models $\mathcal{F}^{MD}$ and $\mathcal{F}^{I}$
present some advantages against models fulfilling the DFS framework.%

\begin{table}[tbp] \centering
\begin{tabular}
[c]{|l|l|l|l|l|l|l|}\hline
${}$ & $\mathcal{M}$ & $\mathcal{M}_{CX}$ & $\mathcal{F}_{owa}$ &
$\mathcal{F}^{MD}$ & $\mathcal{F}^{I}$ & $\mathcal{F}^{A}$\\\hline
\multicolumn{7}{|l|}{{\small Properties derived from the DFS framework}%
}\\\hline
{\small P1. Correct Generalization} & {\small Y} & {\small Y} & {\small Y} &
{\small Y} & {\small Y} & {\small Y}\\\hline
{\small P2. Quantitativity} & {\small Y} & {\small Y} & {\small Y} &
{\small Y} & {\small Y} & {\small Y}\\\hline
{\small P3. Projection quantifiers} & {\small Y} & {\small Y} & {\small Y} &
{\small Y} & {\small Y} & {\small Y}\\\hline
{\small P4. Induced propositional logic} & {\small Y} & {\small Y} &
{\small Y} & {\small Y} & {\small Y} & {\small Y}\\\hline
{\small P6. External negation} & {\small Y} & {\small Y} & {\small Y} &
{\small Y} & {\small Y} & {\small Y}\\\hline
{\small P7. Internal negation} & {\small Y} & {\small Y} & {\small Y} &
{\small N} & {\small finite} & {\small Y}\\\hline
{\small P8. Dualisation} & {\small Y} & {\small Y} & {\small Y} & {\small N} &
{\small finite} & {\small Y}\\\hline
{\small P9. Union/intersection of argument} & {\small Y} & {\small Y} &
{\small Y} & {\small Y} & {\small N} & {\small Y}\\\hline
{\small P10. Coherence with standard quantifiers} & {\small Y} & {\small Y} &
{\small Y} & {\small unary} & {\small unary} & {\small Y}\\\hline
{\small P11. Monotonicity in arguments} & {\small Y} & {\small Y} & {\small Y}
& {\small Y} & {\small Y} & {\small Y}\\\hline
{\small P12. Monotonicity in quantifiers} & {\small Y} & {\small Y} &
{\small Y} & {\small Y} & {\small Y} & {\small Y}\\\hline
{\small P13. Crisp Argument Insertion} & {\small Y} & {\small Y} & {\small Y}
& {\small Y} & {\small Y} & {\small Y}\\\hline
\multicolumn{7}{|l|}{{\small Additional properties}}\\\hline
{\small P14. Continuity in arguments} & {\small Y} & {\small Y} & {\small Y} &
{\small finite} & {\small finite} & {\small finite}\\\hline
{\small P15. Continuity in quantifiers} & {\small Y} & {\small Y} & {\small Y}
& {\small Y} & {\small Y} & {\small Y}\\\hline
{\small P16. Propagation of fuzziness} & {\small Y} & {\small Y} & {\small N}
& {\small N} & {\small N} & {\small N}\\\hline
{\small P17. Fuzzy argument insertion} & {\small N} & {\small Y} & {\small N}
& {\small N} & {\small Y} & {\small Y}\\\hline
\end{tabular}
\caption{Comparison of the behavior of the models against the set of properties in the QFM framework\label{TableSummaryGlockner}}%
\end{table}%

\subsubsection{$\mathcal{M}$ model}

$\mathcal{M}$ model is one of the first models formulated by Gl\"{o}ckner
\cite{Glockner97} and it is also one of the three models for which the author
has provided computational algorithms in \cite[chapter 11]{Glockner06Libro}.
Being an standard DFS $\mathcal{M}$ model fulfills the properties derived from
the DFS framework. Additionally, the $\mathcal{M}$ model is \textit{continuous
in arguments} and \textit{in quantifiers} and fulfills the properties of
\textit{propagation of fuzziness in arguments} and \textit{in quantifiers}.

\subsubsection{$\mathcal{M}_{CX}$ model}

$\mathcal{M}_{CX}$ model is also a standard DFS and another model for which
the author has provided a computational implementation in \cite[chapter
11]{Glockner06Libro}. $\mathcal{M}_{CX}$ is \textit{continuous in arguments}
and \textit{in quantifiers} and fulfills both \textit{fuzziness propagation
properties}. $\mathcal{M}_{CX}$ is considered by Gl\"{o}ckner as a model of
unique properties: it fulfills the property of \textit{fuzzy argument
insertion} \cite[definition 7.82]{Glockner06Libro}, it is specially robust
against modification of membership degrees and generalizes the Sugeno integral
(see \cite[section 7.13]{Glockner06Libro} for more details).

\subsubsection{$\mathcal{F}_{owa}$ model}

$\mathcal{F}_{owa}$ model is the paradigmatic example of an standard DFS that
does not propagate \textit{fuzziness in arguments} or \textit{in}
\textit{quantifiers}. $\mathcal{F}_{owa}$ model is also \textit{continuous in
arguments} and \textit{in quantifiers}. As it fails to fulfill propagation of
fuzziness properties, it is considered as the ideal model for applications in
which an improved discriminative power is necessary \cite[section
8.1]{Glockner06Libro}. $\mathcal{F}_{owa}$ model generalizes Choquet integral.
It is the third model for which a computational implementation has been
provided in \cite[definition 7.82]{Glockner06Libro}.

\subsubsection{$\mathcal{F}^{MD}$ model}

$\mathcal{F}^{MD}$ is the generalization to \textit{QFMs} of the $GD$ model
proposed by Delgado et al. in \cite{Delgado99}, \cite{Sanchez99},
\cite{Delgado00}. $\mathcal{F}^{MD}$ model is not a DFS, failing to fulfill
the \textit{internal negation property}, and as a consequence, the
\textit{dualisation axiom of DFSs (Z3)}. $\mathcal{F}^{MD}$ model is
\textit{continuous in the arguments in the finite case} and also
\textit{continuous in the quantifiers}. $\mathcal{F}^{MD}$ fulfills the
properties of \textit{probabilistic interpretation of quantifiers} and of
\textit{averaging for the identity quantifier} \cite[chapter 3]%
{DiazHermida06Tesis}, that will be reintroduced as one of the criteria for
comparing the behavior of selected QFMs in the following section.
$\mathcal{F}^{MD}$ does not fulfill any of the \textit{propagation of
fuzziness properties}.

\subsubsection{$\mathcal{F}^{I}$ model}

$\mathcal{F}^{I}$ model is the second alpha-cut based model analyzed in
\cite{DiazHermida06Tesis}. $\mathcal{F}^{I}$ model does not fulfill the
\textit{internal joins property (axiom Z4)}, and then fails to be a DFS.
$\mathcal{F}^{I}$ is \textit{continuous in the arguments in finite domains}
and also \textit{continuous in the quantifiers}. $\mathcal{F}^{I}$ model
fulfills the \textit{dualisation property in the finite case}. $\mathcal{F}%
^{I}$ model also fulfills the properties of \textit{probabilistic
interpretation of quantifiers} and \textit{averaging for the identity
quantifier} \cite[chapter 3]{DiazHermida06Tesis}. $\mathcal{F}^{I}$ does not
fulfill \textit{propagation of fuzziness properties}.

\subsubsection{$\mathcal{F}^{A}$ model}

$\mathcal{F}^{A}$ model is, to our knowledge, the unique known non-standard
DFS. The fuzzy operators induced by the model are the \textit{product $tnorm$}
and the \textit{probabilistic sum $tconorm$}, making this model essentially
different of the standard DFSs presented in \cite{Glockner06Libro}. By
definition $\mathcal{F}^{A}$ is a finite model. Moreover, $\mathcal{F}^{A}$ is
\textit{continuous in arguments} and \textit{in quantifiers}, it does not
fulfill fuzziness propagation properties, but it fulfills
\textit{probabilistic interpretation of quantifiers} and \textit{averaging for
the identity quantifier }properties\textit{.}

\section{Some additional criteria to characterize the behavioral differences
of the QFMs}

We have seen that models $\mathcal{F}^{MD}$, $\mathcal{F}^{I}$, $\mathcal{F}%
^{A}$, $\mathcal{M}$, $\mathcal{M}_{CX}$ and $\mathcal{F}_{owa}$ fulfill most
of the adequacy properties that has been presented in \cite{Glockner06Libro}%
.\ If we only took into account properties included in the QFM framework when
selecting a model for an application, we would just choose one of the
best-behaved models (e.g., $\mathcal{F}^{A}$ or $\mathcal{M}_{CX}$) and we
will use them in every possible application of fuzzy quantification.

However, as we will see through this section, properties included in the QFM
framework are not enough for fully understanding the behavioral differences
between the selected models. We will present an analysis that proves that
models here discussed have some strong differences in their behavior. In
addition, an aspect we consider specially relevant is that, from an user
viewpoint, the complexity of the definition of the models and adequacy
properties makes very difficult for a non-specialist in fuzzy quantification
to determine which model should be chosen for a specific application.

Thus, it is essential to establish a set of criteria that help us understand
the behavioral differences between the different models and facilitate the
selection of the more convenient ones for applications. In general, the set of
criteria that we will take into account would not allow us to select `a
perfect model', or even `a preferred one' for every possible application. But
we are convinced they are important to (1) clarify the differences between the
behavior of the QFMs (2) to select or discard QFMs for specific applications
with respect to the behavior we consider more important and (3) to understand
the problems that the selection of a specific model could have for a
particular application.

The following is a summary of the criteria we will consider:

\begin{itemize}
\item \textbf{Linguistic compatibility}. By linguistic compatibility we mean
the fulfillment of the most relevant linguistic properties derived from the
DFS framework. In the summary of the behavior of the main QFMs we have seen
that between the selected models, only DFSs fulfill the main set of properties
that have been established to guarantee an adequate behavior with respect to
the main linguistic expectations.

\item \textbf{Aggregative behavior for low degrees of membership:} aggregative
behavior makes reference to the tendency of a model to confuse one\ `high
degree' membership element with a large quantity of `low degree' membership
elements. It has been one of the main critiques made to the $\sum count$ model
\cite{Zadeh83}, \cite{Zadeh83Collection}.

\item \textbf{Propagation of fuzziness: }Propagation of fuzziness is the main
property used in \cite[section 5.2 and 6.3]{Glockner06Libro} to group the
different classes of standard DFSs \cite[chapters 7 and 8]{Glockner06Libro}.
Basically, models fulfilling the propagation of fuzziness properties
`transfer' fuzziness in inputs to the outputs; that is, they guarantee that
fuzzier inputs and/or fuzzier quantifiers produce fuzzier outputs.

\item \textbf{Identity quantifier: }the `identity semi-fuzzy quantifier' is
defined by means of the identity function $f\left(  x\right)  =\frac{x}{N}$,
$x\in1,\ldots,N$ in the absolute case or by means of $f\left(  x\right)
=x,x\in\left[  0,1\right]  $ in the proportional case. For this semi-fuzzy
quantifier, a linear increase in the number of elements that belong to the
input, produce a linear increase in the output. We could expect that a
reasonable fuzzy counterpart of the identity quantifier should also produce a
linear increase in the output for a linear increase in the input.

\item \textbf{Evaluating quantifiers over `quantified partitions'.} With this
criterion we refer to the behavior of the models when we apply,
simultaneously, a set of quantifiers dividing the quantification universe
(e.g., \textit{`nearly none'}, \textit{`a few'}, \textit{`several'},
\textit{`many'}, \textit{`nearly all'}) to a fuzzy set. That is, how the
degrees of fulfillment of the evaluation of the quantified expressions are
distributed between the labels.

\item \textbf{Fine distinction between objects}. In applications of fuzzy
quantification for ranking generation is generally needed that fuzzy
quantifiers are able to clearly distinguish between objects fulfilling a set
of criteria with different degrees. Criteria to distinguish the QFMs with
respect to their discriminative power are necessary for these applications.
\end{itemize}

\subsection{Linguistic compatibility}

With linguistic compatibility we make reference to the main linguistic
properties presented in \cite[chapter 4 and 6]{Glockner06Libro}. The DFS
framework guarantees that the main linguistic transformations, including
\textit{argument permutations, negation of quantifiers, antonym of
quantifiers, dual of quantifiers, argument insertion, internal meets, etc.
}are transferred from the semi-fuzzy to the fuzzy case.

\subsubsection{Analysis of the models}

\paragraph{Models $\mathcal{M}_{CX}$ and $\mathcal{F}^{A}$}

Being DFSs, both models fulfill all the semantic linguistic properties derived
from the DFS framework. Moreover, these models fulfill the fuzzy argument
insertion property \cite[section 6.8]{Glockner06Libro}, as it can be seen in
\cite[section 7.13]{Glockner06Libro} and in \cite[chapter 3]%
{DiazHermida06Tesis}.

\paragraph{Models $\mathcal{M}$ and $\mathcal{F}_{owa}$}

$\mathcal{M}$ and $\mathcal{F}_{owa}$ models fulfill semantic linguistic
properties derived from the DFS framework, but not fuzzy argument insertion.

\paragraph{Model $\mathcal{F}^{MD}$}

The main difference of $\mathcal{F}^{MD}$ model with respect to DFSs is the
non-fulfillment of the \textit{internal negation property}. This fact impedes
the $\mathcal{F}^{MD}$ model to correctly translate antonym relationships to
the fuzzy case and, as a consequence, duality transformations (see
\cite[chapter 3]{DiazHermida06Tesis}). As an example, failing to fulfill the
\textit{internal negation property} the model cannot guarantee the equivalence
of $\mathcal{F}^{MD}\left(  \mathbf{all}\right)  \left(
\mathbf{hard\_workers},\mathbf{well\_paid}\right)  $ and $\mathcal{F}%
^{MD}\left(  \mathbf{no}\right)  \left(  \mathbf{hard\_workers},\widetilde
{\lnot}\mathbf{well\_paid}\right)  $. In words, results of evaluating
\textit{\textquotedblleft all hard workers are well paid\textquotedblright%
\ }and \textit{\textquotedblleft no hard worker is not well
paid\textquotedblright\ }are different.

$\mathcal{F}^{MD}$ fulfills the strong conservativity property \cite[section
6.7]{Glockner06Libro} that guarantees that conservative semi-fuzzy quantifiers
(i.e., quantifiers fulfilling $Q\left(  Y_{1},Y_{2}\right)  =Q\left(
Y_{1},Y_{1}\cap Y_{2}\right)  $ are correctly translated to the fuzzy case
\cite[chapter 3]{DiazHermida06Tesis}. This property is not fulfilled by any
DFS. However, loosing the internal negation property and as a consequence, the
maintenance of the relationships of the `Aristotelian square' seems more
relevant than the fulfillment of the conservativity property.

\paragraph{Model $\mathcal{F}^{I}$}

Model $\mathcal{F}^{I}$ looses the \textit{internal meets }property. Moreover,
the internal negation property is only fulfilled in the finite case (see
\cite[chapter 3]{DiazHermida06Tesis}).

Loosing the internal meets property $\mathcal{F}^{I}$ model does not guarantee
absolute unary/binary transformations. For example, $\mathcal{F}^{I}\left(
\mathbf{\operatorname{about}\_10}\right)  \left(  \mathbf{hard\_workers}%
,\mathbf{well\_paid}\right)  $ and $\mathcal{F}^{I}\left(
\mathbf{\operatorname{about}\_10}\right)  (\mathbf{hard\_}$ $\mathbf{workers}%
\widetilde{\cap}\mathbf{well\_paid)}$ are not equivalent, and then
\textit{\textquotedblleft about 10 hard workers are well
paid\textquotedblright\ }(evaluated by means of the binary absolute quantifier
\textit{\textquotedblleft about 10\textquotedblright})\ and
\textit{\textquotedblleft about 10 employees are hard workers and are well
paid\textquotedblright\ }(evaluated by means of the unary version of the
absolute quantifier \textit{`about 10'\ }and the induced $tnorm$ of the model
used to compute the intersection of \ \textit{`hard workers'} and
\textit{`well paid'}) will not produce the same results.

\subsection{Aggregative behavior for low degrees of membership}

Aggregative behavior for low degrees of membership is one of the main
critiques that has been made to the Zadeh's $\sum count$ model \cite{Yager93}%
,\cite[section A.3]{Glockner06Libro}, \cite{Barro02}. The intuition around
aggregative behavior is that in evaluating quantified expressions, a large
amount of elements fulfilling a property with `low degree' of membership
should not be confused with a small amount of elements fulfilling a property
with `high degree' of membership. In the case of the Zadeh's model is easy to
understand the meaning of aggregative behavior as:%

\[
\sum count\left(  \exists\right)  \left(  \left\{  0.01/e_{1},\ldots
,0.01/e_{100}\right\}  \right)  =\sum count\left(  \exists\right)  \left(
\left\{  1/e_{1},0/e_{2},\ldots,0/e_{100}\right\}  \right)  =1
\]
in words, \textit{`exist one tall person'} can be fulfilled if there exists
exactly \textit{`one tall person'}, or if there exist 100 people being
\textit{`0.01 tall'}.

Although intuitions against aggregative behavior seem clear, giving up models
presenting aggregative behavior will force us to discard non-standard DFS
$\mathcal{F}$ associated to \textit{archimedean} $tconorms$\footnote{For an
\textit{archimedean }tconorm, $\lim_{n\longrightarrow\infty}\vee\left(
c/e_{1},\ldots,c/e_{n}\right)  =1$.\ Every continuous tconorm such that
$\vee\left(  x,x\right)  >x,x\in\left(  0,1\right)  $ is \textit{archimedean.}%
}. Being $\mathcal{F}\left(  \exists\right)  $ equal to (\cite[Theorem
4.61]{Glockner06Libro}):%

\[
\mathcal{F}\left(  \exists\right)  \left(  X\right)  =\sup\left\{
\widetilde{\vee}_{i=1}^{m}\left(  a_{i}\right)  :A=\left\{  a_{1},\ldots
,a_{m}\right\}  \in\mathcal{P}\left(  E\right)  \text{ }finite,a_{i}\neq
a_{j}\text{ }if\text{ }i\neq j\right\}
\]
for all $X\in\widetilde{\mathcal{P}}\left(  E\right)  $, then $\mathcal{F}%
\left(  \exists\right)  \left(  X\right)  $ will always present aggregative
behavior for every non-standard DFS associated to an \textit{archimedean}
$tconorm$ $\vee$. \textit{Archimedean} $tconorms$ are a very relevant class of
$tconorm$ operators, including most of the common examples of $tconorm$ operators.

To the best of our knowledge, a clear definition of aggregative behavior has
not been presented in the literature, that has limited itself to present
examples with existential quantifiers and/or with proportional quantifiers
representing small proportions (e.g., \textit{`about 10\%'}). In this
discussion, we will limit us to consider aggregative behavior for existential
quantifiers, as it is enough to characterize the models we are considering.

\subsubsection{Analysis of the models}

\paragraph{Model $\mathcal{F}^{A}$}

$\mathcal{F}^{A}$ model presents aggregative behavior as a consequence of
inducing the non-standard probabilistic sum \textit{$tconorm$ }$\widetilde
{\vee}\left(  a,b\right)  =a+b-ab$. For the $\mathcal{F}^{A}$ model:%
\[
\mathcal{F}\left(  \exists\right)  \left(  X\right)  =\widetilde{\vee}_{e\in
E}\mu_{X}\left(  e\right)
\]

Moreover, $\mathcal{F}^{A}$ model tends to the Zadeh's Sigma-count model when
the size of the referential $E$ tends to infinite \cite{DiazHermida10Arxiv};
that is:%
\[
\lim_{\left\vert E\right\vert \rightarrow\infty}\mathcal{F}^{A}\left(
Q\right)  \left(  X\right)  =\mu_{Q}\left(  \frac{\sum_{e\in E}\mu_{X}\left(
e\right)  }{\left\vert E\right\vert }\right)
\]

In this way, $\mathcal{F}^{A}$ shares the critiques of aggregative behavior
that has been made to the Zadeh's model for large referential sets.

\paragraph{Models $\mathcal{M}$, $\mathcal{M}_{CX}$, $\mathcal{F}_{owa}$,
$\mathcal{F}^{MD}$ and $\mathcal{F}^{I}$}

None of the rest of the models show aggregative behavior. For all of them,
$\mathcal{F}\left(  \exists\right)  \left(  X\right)  =\sup\left\{  \mu
_{X}\left(  e\right)  :e\in E\right\}  $, $E$ finite. We will give some
intuitions about the reasons for which these models do not present aggregative behavior.

Model $\mathcal{M}_{CX}$ has been proved to be extremely stable. In
\cite[section 7.12]{Glockner06Libro} it is proved that a change in the
arguments that does not exceed a given $\Delta$ will not change the result of
the quantifier by more than $\Delta$. Then, $\mathcal{F}\left(  Q\right)
\left(  \varnothing\right)  $ and $\mathcal{F}\left(  Q\right)  \left(
\left\{  c/e_{1},c/e_{2},\ldots,c/e_{N}\right\}  \right)  $, with $c$ `small',
will produce approximately the same results.

With respect to models $\mathcal{M}$, $\mathcal{F}_{owa}$, $\mathcal{F}^{MD}$
and $\mathcal{F}^{I}$ we should take into account that all of their
definitions are made by using an integration process over the alpha-cuts or
the three-valued cuts of the argument sets.

In the case of alpha cuts, only alpha cuts in the integration interval
$\left(  0,c\right]  $ could be altered by modifications in degrees of
membership of elements with membership degree $\mu_{X}\left(  e\right)  \leq
c\ $that are maintained in $\left(  0,c\right]  $. In the case of three-valued
cuts, only the integration interval $\left[  1-2c,1\right]  $ could be altered
by modifications in degrees of membership in the same interval. As the results
of the integral do not change out of the integration range, effects of
modifications lower or equal than $c$ will be limited to $c$ (in the case of
alpha-cuts) or $2c$ in the case of three-valued cuts\footnote{It can be proved
that differences in the integration process for $\mathcal{M}$ and
$\mathcal{F}_{Ch}$ are also limited to $c$, but we are only interested in
giving an intuitive explanation of the reasons for which these models do not
present aggregative behaviour.}.

\subsection{Propagation of fuzziness\label{SectionPropagationFuzziness}}

Propagation of fuzziness is related with the transmission of imprecision from
the inputs (arguments and quantifiers) to the outputs (results of evaluating
quantified expressions). We will reproduce the main definitions in
\cite[section 5.2 and 6.3]{Glockner06Libro}.

Let be $\preceq_{c}$ a partial order in $\mathbf{I}\times\mathbf{I}$ defined
as%
\[
x\preceq_{c}y\Leftrightarrow y\leq x\leq\frac{1}{2}\text{ or }\frac{1}{2}\leq
x\leq y
\]
for $x,y\in\mathbf{I}$.

$\preceq_{c}$ can be extended to fuzzy sets, semi-fuzzy quantifiers and fuzzy
quantifiers in the following way:%

\begin{align*}
X  &  \preceq_{c}X^{\prime}\Leftrightarrow\mu_{X}\left(  e\right)  \preceq
_{c}\mu_{X^{\prime}}\left(  e\right)  ,\text{ for all }e\in E\\
Q  &  \preceq_{c}Q^{\prime}\Leftrightarrow Q\left(  Y_{1},\ldots,Y_{n}\right)
\preceq_{c}Q^{\prime}\left(  Y_{1},\ldots,Y_{n}\right)  ,\text{ for all }%
Y_{1},\ldots,Y_{n}\in\mathcal{P}\left(  E\right) \\
\widetilde{Q}  &  \preceq_{c}\widetilde{Q^{\prime}}\Leftrightarrow
\widetilde{Q}\left(  X_{1},\ldots,X_{n}\right)  \preceq_{c}\widetilde
{Q^{\prime}}\left(  X_{1},\ldots,X_{n}\right)  ,\text{ for all }X_{1}%
,\ldots,X_{n}\in\widetilde{\mathcal{P}}\left(  E\right)
\end{align*}

\begin{defn}
\cite[section 6.3]{Glockner06Libro}. Let a $QFM$ $\mathcal{F}$ be given.
\end{defn}

a. We say that $\mathcal{F}$ propagates fuzziness in arguments if the
following property $Q\preceq_{c}Q^{\prime}$is satisfied for all $Q:\mathcal{P}%
\left(  E\right)  ^{n}\rightarrow\mathbf{I}$ and $X_{1},\ldots,X_{n}%
,X_{1}^{\prime},\ldots,X_{n}^{\prime}\in\widetilde{\mathcal{P}}\left(
E\right)  $. If $X_{i}\preceq_{c}X_{i}^{\prime}$ for all $i=1,\ldots,n$ then
$\mathcal{F}\left(  Q\right)  \left(  X_{1},\ldots,X_{n}\right)  \preceq
_{c}\mathcal{F}\left(  Q\right)  \left(  X_{1}^{\prime},\ldots,X_{n}^{\prime
}\right)  $.

b. We say that $\mathcal{F}$ propagates fuzziness in quantifiers if
$\mathcal{F}\left(  Q\right)  \preceq_{c}\mathcal{F}\left(  Q^{\prime}\right)
$ whenever $Q\preceq_{c}Q^{\prime}$.

\smallskip Propagation of fuzziness in arguments and in quantifiers is
considered as optional but really convenient in \cite[section 6.3]%
{Glockner06Libro}. Intuitively, from an user point of view, fuzzier inputs or
fuzzier quantifiers should not produce more specific outputs.

Although both propagation of fuzziness properties seem natural, we should note
that most basic \textit{$tnorms$} and \textit{$tconorms$} do not fulfill
propagation of fuzziness properties (e.g. product \textit{$tnorm$} and
probabilistic sum \textit{$tconorm$}). This fact is relevant, as every DFS
embeds basic logic operators \cite[section 3.4]{Glockner06Libro}. Moreover,
fulfillment of propagation of fuzziness properties have strong negative
consequences for the ranking of objects (see section
\ref{SectionRankingGeneration}).

\subsubsection{Analysis of the models}

\paragraph{Models $\mathcal{M}$, $\mathcal{M}_{CX}$}

Models $\mathcal{M}$ and $\mathcal{M}_{CX}$ are the paradigmatic examples of
standard DFSs fulfilling propagation of fuzziness properties (see
\cite[chapter 7]{Glockner06Libro}. Using $\mathcal{M}$ and $\mathcal{M}_{CX}$
assure that when presented with fuzzier inputs or quantifiers, we will always
obtain fuzzier outputs.

\paragraph{Models $\mathcal{F}_{owa},\mathcal{F}^{A},\mathcal{F}^{MD}$ and
$\mathcal{F}^{I}$}

Model $\mathcal{F}_{owa}$ is the paradigmatic example of an standard DFSs that
does not fulfill both propagation of fuzziness properties.

$\mathcal{F}^{A}$ model does not fulfill propagation of fuzziness in
arguments, as it is not fulfilled by the induced product \textit{$tnorm$} and
the induced probabilistic sum \textit{$tconorm$} of the model (see
\cite[chapter 3]{DiazHermida06Tesis}) and it is easy to find counterexamples
for propagation of fuzziness in quantifiers. $\mathcal{F}^{MD}$ and
$\mathcal{F}^{I}$ do not fulfill the property of propagation of fuzziness in
arguments (see \cite[chapter 3]{DiazHermida06Tesis}) and it is also trivial to
find counterexamples for the property of propagation of fuzziness in quantifiers.

\subsection{Identity quantifier: as many as
possible\label{IdentitiyQuantifier}}

First, we will define the \textbf{identity} semi-fuzzy quantifier. We will
limit us to the proportional case:

\begin{defn}
The unary semi-fuzzy quantifier \textbf{identity}$:\mathcal{P}\left(
E\right)  \rightarrow\mathbf{I}$\textit{ is defined as}%
\[
\mathbf{identity}\left(  Y\right)  =\frac{\left\vert Y\right\vert }{\left\vert
E\right\vert },Y\in\mathcal{P}\left(  E\right)
\]

\end{defn}

For the \textbf{identity} semi-fuzzy quantifier, adding one element increments
the result in $\frac{1}{m}$. Thus, the increase in the output obtained with
the addition of elements to the argument set is linear, making possible to
interpret $\mathbf{identity}\left(  Y\right)  $ as \textit{`as many as
possible'\ }or \textit{`the more the better'}. In other way, the identity
semi-fuzzy quantifier measures the relative weight of the input set $Y$ with
respect to the referential set $E$. That is, $\mathbf{identity}\left(
Y\right)  =\left\vert Y\right\vert /\left\vert E\right\vert $.

A plausible fuzzy counterpart of the identity quantifier should also produce a
linear increase in the output for a linear increase in the input.

\begin{defn}
\cite[chapter 3]{DiazHermida06Tesis} We will say that a QFM $\mathcal{F}$
fulfills the average property for the identity quantifier if:%
\[
\mathcal{F}\left(  \mathbf{identity}\right)  \left(  X\right)  =\frac{1}%
{m}\sum_{j=1}^{m}\mu_{X}\left(  e_{j}\right)
\]

\end{defn}

As a result of the fulfillment of the average property for the identity
quantifier, the improvement obtained in $\mathcal{F}\left(  \mathbf{identity}%
\right)  \left(  X\right)  $ is linear with respect to the increase of the
membership grades of the argument fuzzy set. This property allows us to
enquire if this intuition is translated to the fuzzy case, assuring that in
the fuzzy case we will obtain a measure of the relative weight of
$X\in\widetilde{\mathcal{P}}\left(  E\right)  $ with respect to $E$.

\subsubsection{Analysis of the models}

\paragraph{Models $\mathcal{F}_{owa}$, $\mathcal{F}^{MD}$, $\mathcal{F}^{I}$
and $\mathcal{F}^{A}$}

Model $\mathcal{F}_{owa}$ \cite[chapter 8]{Glockner06Libro}, and models
$\mathcal{F}^{MD}$, $\mathcal{F}^{I}$ \cite{DiazHermida06Tesis} generalize the
OWA\ approach, and then they trivially fulfill the property of averaging for
the identity quantifier. Model $\mathcal{F}^{A}$ also fulfills this property
\cite{DiazHermida06Tesis},\cite{DiazHermida10Arxiv}.

\paragraph{Models $\mathcal{M}$ and $\mathcal{M}_{CX}$}

Models $\mathcal{M}$ and $\mathcal{M}_{CX}$ do not fulfill this property, as a
direct consequence of fulfilling the propagation of fuzziness in the
arguments. For $\mathcal{M}$ and $\mathcal{M}_{CX}$ models, if $\mathcal{M}%
\left(  X\right)  =a$ or $\mathcal{M}_{CX}\left(  X\right)  =a$, $a\geq0.5$,
then $\mathcal{M}\left(  X^{\prime}\right)  ,\mathcal{M}_{CX}\left(
X^{\prime}\right)  \in\left[  0.5,a\right]  $ for $X^{\prime}\preceq_{c}X$
($X^{\prime}$ fuzzier than $X$). More clearly:%
\begin{align*}
&  \mathcal{M}\left(  \mathbf{identity}\right)  \left(  \left(  \left\{
1/e_{1},1/e_{2},0/e_{3},0/e_{4}\right\}  \right)  \right)  =\mathcal{M}%
_{CX}\left(  \mathbf{identity}\right)  \left(  \left(  \left\{  1/e_{1}%
,1/e_{2},0/e_{3},0/e_{4}\right\}  \right)  \right)  \\
&  =\mathcal{M}_{CX}\left(  \mathbf{identity}\right)  \left(  \left(  \left\{
0.5/e_{1},0.5/e_{2},0.5/e_{3},0.5/e_{4}\right\}  \right)  \right)  \\
&  =\mathcal{M}_{CX}\left(  \mathbf{identity}\right)  \left(  \left(  \left\{
1/e_{1},1/e_{2},0.5/e_{3},0.5/e_{4}\right\}  \right)  \right)  \\
&  =\mathcal{M}_{CX}\left(  \mathbf{identity}\right)  \left(  \left(  \left\{
0.5/e_{1},0.5/e_{2},0/e_{3},0/e_{4}\right\}  \right)  \right)  \\
&  =0.5
\end{align*}
that is, as $\mathcal{M}\left(  \mathbf{identity}\right)  \left(  \left(
\left\{  1/e_{1},1/e_{2},0/e_{3},0/e_{4}\right\}  \right)  \right)  =0.5$ then
for every possible $X^{\prime}$ such that $X^{\prime}\preceq_{c}X$ the result
will be at least as fuzzier as $0.5$, but as $0.5$ is the fuzzier possible
output, $\mathcal{M}\left(  \mathbf{identity}\right)  \left(  X^{\prime
}\right)  =0.5$.

\begin{figure}
[ptb]
\begin{center}
   \includegraphics[width=0.60\columnwidth]{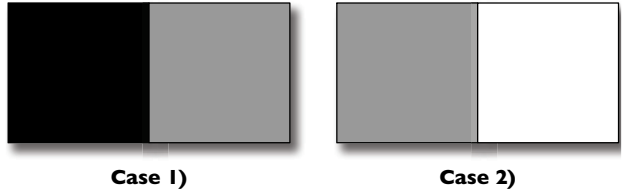}
\caption{Indiscernible situations for the identity quantifier}%
\label{FigGrises}%
\end{center}
\end{figure}

In figure \ref{FigGrises} we show a graphic representation of this behavior.
Although we would expect a high degree of fulfillment for the identity
quantifier in case 1 and a low degree in case 2, results of applying
$\mathcal{M}$ or $\mathcal{M}_{CX}$ to the identity quantifier for both inputs
is $0.5$ for every intermediate case between case 1) and case 2).%

\subsection{Evaluating quantifiers over `quantified partitions'}

\begin{figure}
[ptb]
\begin{center}
   \includegraphics[width=0.60\columnwidth]{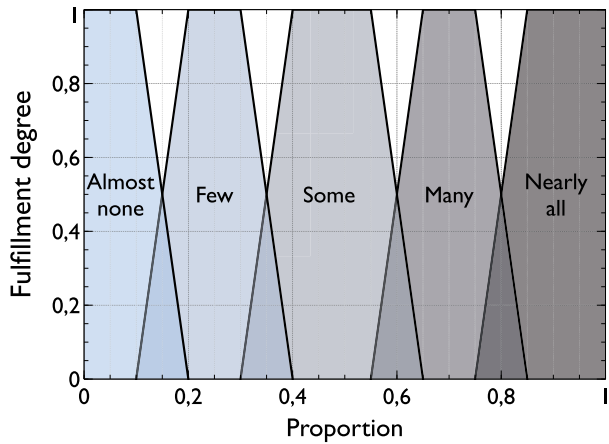}
\caption{Partition of the quantified universe.}%
\label{FigQuantifiedPartition}%
\end{center}
\end{figure}

In this section we will analyze the behavior of the models when we
simultaneously evaluate a set of fuzzy quantifiers associated to a `quantified
partition' of the quantification universe. Let us consider the set of
quantification labels presented in figure \ref{FigQuantifiedPartition}.%

For reasons we will see later on, we will restrict us to a set of labels such
that $\mu_{Q_{i}}\left(  x\right)  +\mu_{Q_{i+1}}\left(  x\right)  =1$ for
some $i$. In any case, this is a very common way of dividing the reference
universe in practical applications. We will refer to quantified partitions
fulfilling this property as `Ruspini quantified partitions'.

When we consider the simultaneous evaluation of a set of quantifiers defined
by means of a quantified partition, behavior of standard DFSs ($\mathcal{M}$,
$\mathcal{M}_{CX}$ $\mathcal{F}_{owa}$) and QFMs ($\mathcal{F}^{A}$,
$\mathcal{F}^{I}$ and $\mathcal{F}^{MD}$) present strong differences. For some
situations, models $\mathcal{M}$, $\mathcal{M}_{CX}$ and $\mathcal{F}_{owa}$
tend to produce the $0.5$ output for every quantifier in the partition.
Contrasting with these behavior, models $\mathcal{F}^{A}$, $\mathcal{F}^{I}$
and $\mathcal{F}^{MD}$ guarantee that the sum of the evaluation results equals
$1$, producing `a distribution' of the truth between the set of quantified labels.

\subsubsection{Analysis of the models}

\paragraph{Models $\mathcal{M}$, $\mathcal{M}_{CX}$ and $\mathcal{F}_{owa}$}

As a consequence of being based in trivalued cuts models $\mathcal{M}$,
$\mathcal{M}_{CX}$ and $\mathcal{F}_{owa}$ present a tendency to produce $0.5$
evaluation results for some situations. Let us consider a fuzzy set $X$ such
that%
\[
X=\left\{  0.5/e_{1},0.5/e_{2},\ldots,0.5/e_{m}\right\}
\]
then, if for a semi-fuzzy quantifier $Q_{i}$ is fulfilled that there exist
$r,j$ such that $q_{i}\left(  r\right)  =1$ and $q_{i}\left(  j\right)  =0$
(note that every quantifier in figure \ref{FigQuantifiedPartition} fulfill
this property) then
\[
\mathcal{M}\left(  X\right)  =\mathcal{M}_{CX}\left(  X\right)  =\mathcal{F}%
_{owa}\left(  X\right)  =0.5
\]
as it can be easily checked.

This behavior is independent of the granularity of the partition. That is,
finer partitions will continue to produce a $0.5$ output in each situation in
which we could find some $i,j$ such that $q_{i}\left(  i\right)  =1$ and
$q_{i}\left(  j\right)  =0$.

\paragraph{Model $\mathcal{F}^{A}$}

Before presenting the behavior of the $\mathcal{F}^{A}$ model, we need to
introduce some definitions to precise the meaning of a `Ruspini quantified partition'.

\begin{defn}
We will say that a set of semi-fuzzy quantifiers $Q_{1},\ldots,Q_{r}%
:\mathcal{P}^{n}\left(  E\right)  \rightarrow\mathbf{I}$ forms a Ruspini
partition of the quantification universe if for all $Y_{1},\ldots,Y_{n}%
\in\mathcal{P}\left(  E\right)  $ it holds that%
\[
Q_{1}\left(  Y_{1},\ldots,Y_{n}\right)  +\ldots+Q_{r}\left(  Y_{1}%
,\ldots,Y_{n}\right)  =1
\]

\end{defn}

\begin{exmp}
The next set of quantifiers form a Ruspini partition of the quantification universe:
\end{exmp}%

\[
Q_{i}\left(  Y_{1},Y_{2}\right)  =\left\{
\begin{array}
[c]{cc}%
label_{i}\left(  \frac{\left\vert Y_{1}\cap Y_{2}\right\vert }{\left\vert
Y_{1}\right\vert }\right)   & Y_{1}\neq\varnothing\\
\frac{1}{5} & Y_{1}=\varnothing
\end{array}
\right.
\]
where the `i-th' fuzzy number in the partition is represented by $label_{i}$
represents. This set of fuzzy numbers form a Ruspini partition of the
quantification universe as $\sum_{i}Q_{i}\left(  Y_{1},Y_{2}\right)  =1$ for
all $Y_{1},Y_{2}\in\mathcal{P}\left(  E\right)  $.

\begin{defn}
\cite[chapter 3]{DiazHermida06Tesis}We will say that a QFM $\mathcal{F}$
fulfills the property of probabilistic interpretation of quantifiers if for
all the Ruspini partitions of the quantification universe $Q_{1},\ldots
,Q_{r}:\mathcal{P}\left(  E\right)  ^{n}\rightarrow\mathbf{I}$ it holds that
\[
\mathcal{F}\left(  Q_{1}\right)  \left(  X_{1},\ldots,X_{n}\right)
+\ldots+\mathcal{F}\left(  Q_{r}\right)  \left(  X_{1},\ldots,X_{n}\right)
=1
\]

\end{defn}

This property is very interesting because it allow us to interpret the result
of evaluating fuzzy quantified expressions as probabilities over the labels
related to the quantifiers\footnote{In \cite{Lawry01-Computing} a
probabilistic interpretation of quantifiers is also used under the label
semantics interpretation of fuzzy sets.
\par
{}}. $\mathcal{F}^{A},\mathcal{F}^{MD}$ and $\mathcal{F}^{I}$ models fulfill
this property \cite[chapter three]{DiazHermida06Tesis},
\cite{DiazHermida10Arxiv}. Thus, we can interpret that these models tend to
distribute the truth between the set of labels of the partition, assuring that
the sum of the evaluation results associated to each label adds to $1$.

In addition, in \cite{DiazHermida10Arxiv} it has been proved that for unary
quantifiers the $\mathcal{F}^{A}$ model tends to the Zadeh's Sigma-count model
when the size of the referential $E$ tends to infinite; that is:%
\[
\lim_{\left\vert E\right\vert \rightarrow\infty}\mathcal{F}^{A}\left(
Q\right)  \left(  X\right)  =\mu_{Q}\left(  \frac{\sum_{e\in E}\mu_{X}\left(
e\right)  }{\left\vert E\right\vert }\right)
\]

In this way, for a big $\left\vert E\right\vert $ we have $\mathcal{F}%
^{A}\left(  Q\right)  \left(  X\right)  \approx\mu_{Q}\left(  \frac{\sum_{e\in
E}\mu_{X}\left(  e\right)  }{\left\vert E\right\vert }\right)  $. As
$\frac{\sum_{e\in E}\mu_{X}\left(  e\right)  }{\left\vert E\right\vert }$ is a
punctual value, when we apply this result to a Ruspini quantified partition
like the one presented in figure \ref{FigQuantifiedPartition}, the weights of
the evaluation of the quantified expressions tend to concentrate themselves in
one quantified label $q_{i}$ (being $\mathcal{F}^{A}\left(  Q_{i}\right)
\left(  X\right)  \approx1$) or two contiguous ones $q_{i},q_{i+1}$ (being
$\mathcal{F}^{A}\left(  Q_{i}\right)  +\mathcal{F}^{A}\left(  Q_{i+1}\right)
\approx1$).

\paragraph{Models $\mathcal{F}^{MD}$ and $\mathcal{F}^{I}$}

Models $\mathcal{F}^{MD}$ and $\mathcal{F}^{I}$ also fulfill the property of
\textit{probabilistic interpretation of quantifiers}. Hence, the result of
evaluating a set of quantifiers $Q_{1},\ldots,Q_{r}$ forming a Ruspini
quantified partition can be interpreted as a probability defined over the
quantified labels of the quantifiers. Thus, we can interpret that these models
tend to distribute the weight of evaluating quantified sentences over the set
of labels used to define the fuzzy quantifiers.

Moreover, the following result proves that in the unary case, for a
`perfectly' distributed fuzzy set, models $\mathcal{F}^{MD}$ and
$\mathcal{F}^{I}$ tend to assign to each quantifier a probability weight
proportional to its area. Let us define an equispaced fuzzy set over $\left[
0,1\right]  $ as:%

\[
\mu_{X}\left(  a_{0}\right)  =0<\mu_{X}\left(  a_{1}\right)  =h<\mu_{X}\left(
a_{2}\right)  =2h,\ldots,\mu_{X}\left(  a_{m}\right)  =1
\]
Then, if we restrict ourselves to piecewise continuous functions, it is
fulfilled that:%

\[
\lim_{m\rightarrow\infty}\int_{0}^{1}Q\left(  X_{\geq\alpha}\right)
d\alpha=\lim_{m\rightarrow\infty}\mu_{Q}\left(  \frac{1}{m}\right)  h+\mu
_{Q}\left(  \frac{2}{m}\right)  h+\ldots+\mu_{Q}\left(  \frac{m-1}{m}\right)
h=\int_{0}^{1}\mu_{Q}\left(  x\right)  dx
\]
, as we are simply computing the area of the quantifier.

As a consequence, when we evaluate a set of unary quantifiers $Q_{1}%
,\ldots,Q_{r}:\mathcal{P}\left(  E\right)  \rightarrow\mathbf{I}$ over a fuzzy
set following an identity function ($\mu_{X}\left(  e_{i}\right)  =\frac{i}%
{m}$) we obtain:%
\[
\mathcal{F}^{MD}\left(  Q_{i}\right)  \left(  X\right)  =\mathcal{F}%
^{I}\left(  Q_{i}\right)  \left(  X\right)  \approx area\left(  \mu_{Q_{i}%
}\right)
\]

This property is related to the probabilistic alpha-cut interpretation of
models $\mathcal{F}^{MD}$ and $\mathcal{F}^{I}$. In this interpretation, if
membership values of $X$ are perfectly distributed, then `weights' of the
alpha cuts are perfectly distributed over the quantification universe. In this
way, quantifiers of greater areas tend to `collect' more weight than
quantifiers with smaller areas. This also means that, for `finer' quantifier
partitions, weights tend to be more distributed between the quantifiers of the partition.

\subsection{Fine distinction between objects\label{SectionRankingGeneration}}

In applications of fuzzy quantifiers for ranking generation, we generally have
a set of objects $o_{1},\ldots,o_{N}$ for which the fuzzy fulfillment of a set
of criteria $p_{1},\ldots,p_{m}$ is known $X^{o_{i}}=\left\{  \mu_{X^{i}%
}\left(  p_{1}\right)  /p_{1},\ldots,\mu_{X^{i}}\left(  p_{m}\right)
/p_{m}\right\}  $, where $\mu_{X^{i}}\left(  p_{j}\right)  /p_{j}$ represents
the fulfillment of the criteria $p_{j}$ by the object $o_{i}$. Additionally,
we generally have a set of weights $W=\left\{  \mu_{W}\left(  p_{1}\right)
/p_{1},\ldots,\mu_{W}\left(  p_{m}\right)  /p_{m}\right\}  $ indicating the
relative importance of the criteria $p_{1},\ldots,p_{m}$.

Fuzzy quantification can be used to generate a ranking by means of the
assignment of a weight to each object, computed using an unary proportional
quantified expression, $r^{o_{i}}=\widetilde{Q}\left(  X^{o_{i}}\right)  $
when a vector of weights is not involved, or computed using a binary
proportional quantified expression $r^{o_{i}}=\widetilde{Q}\left(  W,X^{o_{i}%
}\right)  $ when there exists a vector of weights $W$ to indicate the relative
importance of each criteria. Hence, when we compute $r^{o_{i}}$ for each
$i=1,\ldots,N$, we can rank each object with respect to \textit{`how
}$\widetilde{Q}$' criteria it fulfills (e.g., for $\widetilde{Q}%
=\mathbf{many}$, \textit{`how many'}).

Fuzzy quantifiers seem specially convenient for ranking applications. As
$r^{o_{i}}$ indicates `how good' is the object $i$ in fulfilling
`$\widetilde{Q}$ criteria'. We can easily adjust the quantifiers to prioritize
objects fulfilling \textit{`most of the criteria'}, \textit{`some of them'},
\textit{`a least 10'}, etc.

Ranking applications usually demand a great discriminative power between
objects. In general, we should expect that even small variations in the inputs
would produce some effect in the outputs. In order to analyze the
discriminative power of QFMs, we will need some definitions:

\begin{defn}
Let $h\left(  x\right)  :\left[  0,1\right]  \rightarrow\mathbf{I}$ an
strictly increasing continuous mapping; i.e., $h\left(  x\right)  >h\left(
y\right)  $ for every $x>y$.\ We define the unary and binary semi-fuzzy
quantifiers $Q_{h}:\mathcal{P}\left(  E\right)  \rightarrow\mathbf{I}$ and
$Q_{h}:\mathcal{P}\left(  E\right)  ^{2}\rightarrow\mathbf{I}$ as%
\begin{align*}
Q_{h}\left(  Y\right)   &  =h\left(  \left\vert Y\right\vert \right)
,Y\in\mathcal{P}\left(  E\right) \\
Q_{h}\left(  Y_{1},Y_{2}\right)   &  =\left\{
\begin{array}
[c]{cc}%
h\left(  \frac{\left\vert Y_{1}\cap Y_{2}\right\vert }{\left\vert
Y_{1}\right\vert }\right)  & X_{1}\neq\varnothing\\
1 & X_{1}=\varnothing
\end{array}
\right.
\end{align*}

\end{defn}

For assuring the discriminative power of QFMs, we will require that in the
case of unary quantifiers, any increase in the fulfillment of a criteria will
increase $\mathcal{F}\left(  Q_{h}\right)  $. In the binary case, we will
require that any increase in the fulfillment of a criteria associated with a
strictly positive weight will also increase $\mathcal{F}\left(  Q_{h}\right)
$. That is, as $h$ is strictly increasing, we expect that an increase in the
values of the inputs is translated into an increase in the result of the evaluation.

\begin{defn}
Let us consider $X_{1},X_{2}$ such that $\mu_{X_{1}}\left(  e_{i}\right)
=\mu_{X_{2}}\left(  e_{i}\right)  ,i\neq j$, $\mu_{X_{1}}\left(  e_{i}\right)
<\mu_{X_{2}}\left(  e_{i}\right)  ,i=j$. We say that a QFM $\mathcal{F}$
fulfills the property o\textit{f discriminative ranking generation} for unary
quantifiers if:%
\[
\mathcal{F}\left(  Q_{h}\right)  \left(  X_{2}\right)  >\mathcal{F}\left(
Q_{h}\right)  \left(  X_{1}\right)
\]
for $h\left(  x\right)  $ strictly increasing.
\end{defn}

\begin{defn}
Let us consider $W,X_{1},X_{2}$ such that $\mu_{X_{1}}\left(  e_{i}\right)
=\mu_{X_{2}}\left(  e_{i}\right)  ,i\neq j$, $\mu_{X_{1}}\left(  e_{i}\right)
<\mu_{X_{2}}\left(  e_{i}\right)  ,\mu_{W}\left(  i\right)  >0,i=j$. We say
that a QFM $\mathcal{F}$ fulfills the property of\textit{\ discriminative
ranking generation} for binary quantifiers if it fulfills:%
\[
\mathcal{F}\left(  Q_{h}\right)  \left(  W,X_{2}\right)  >\mathcal{F}\left(
Q_{h}\right)  \left(  W,X_{1}\right)
\]
for $h\left(  x\right)  $ strictly increasing.
\end{defn}

\subsubsection{Analysis of the models}

\paragraph{Models $\mathcal{M}$ and $\mathcal{M}_{CX}$}

Fulfillment of propagation of fuzziness properties makes $\mathcal{M}$ and
$\mathcal{M}_{CX}$ very inconvenient for ranking applications. Examples
presented in section \ref{IdentitiyQuantifier} have shown that these models
are piecewise constant, and that they are not able of differentiating between
really large regions of the input space. As a consequence, these models are
incapable of making fine distinction between objects.

\paragraph{Model $\mathcal{F}_{owa}$}

Model $\mathcal{F}_{owa}$ have been presented in \cite[chapter 8]%
{Glockner06Libro} as the paradigmatic example of a standard DFS non-fulfilling
the properties of propagation of fuzziness. Thus, the author consider the
$\mathcal{F}_{owa}$ model convenient for applications needing an `enhanced
discriminatory force'.

As $\mathcal{F}_{owa}$ model generalizes OWA, it adequately deals with the
fine distinction between objects in the unary case. But in the binary case,
$\mathcal{F}_{owa}$ is piecewise constant, as it proves the following example:%

\begin{align*}
&  \mathcal{F}_{owa}\left(  Q_{id}\right)  \left(  \left\{  1/e_{1}%
,1/e_{2},0.5/e_{3},0.5/e_{4}\right\}  ,\left\{  1/e_{1},1/e_{2},0/e_{3}%
,0/e_{4}\right\}  \right) \\
&  =0.75\\
&  =\mathcal{F}_{owa}\left(  Q_{id}\right)  \left(  \left\{  1/e_{1}%
,1/e_{2},0.5/e_{3},0.5/e_{4}\right\}  ,\left\{  1/e_{1}1/e_{2},0.5/e_{3}%
,0.5/e_{4}\right\}  \right)
\end{align*}
In previous example, for $0.5$ weights of $e_{3},e_{4}$, we can modify object
fulfillment in the $\left[  0,0.5\right]  $ range without obtaining any
difference in the output.

\paragraph{Model $\mathcal{F}^{MD}$}

A similar problem happens with the $\mathcal{F}^{MD}$ model. As the
$\mathcal{F}^{MD}$ fulfills the strong conservativity property (see
\cite[chapter 3]{DiazHermida06Tesis}) we have
\[
\mathcal{F}^{MD}\left(  Q_{id}\right)  \left(  W,X^{o_{i}}\right)
=\mathcal{F}^{MD}\left(  Q_{id}\right)  \left(  W,W\widetilde{\cap}X^{o_{i}%
}\right)
\]
and then,
\begin{align*}
&  \mathcal{F}^{MD}\left(  Q_{id}\right)  \left(  \left\{  1/e_{1}%
,1/e_{2},0.5/e_{3},0.5/e_{4}\right\}  ,\left\{  1/e_{1},1/e_{2},1/e_{3}%
,1/e_{4}\right\}  \right) \\
&  =\mathcal{F}^{MD}\left(  Q_{id}\right)  \left(  \left\{  1/e_{1}%
,1/e_{2},0.5/e_{3},0.5/e_{4}\right\}  ,\left\{  1/e_{1},1/e_{2},0.5/e_{3}%
,0.5/e_{4}\right\}  \right) \\
&  =1
\end{align*}

\paragraph{Model $\mathcal{F}^{I}$}

Model $\mathcal{F}^{I}$ fulfills the property of discriminative ranking
generation. The proof is shown in the Apendix.

\paragraph{Model $\mathcal{F}^{A}$}

The $\mathcal{F}^{A}$ model also fulfills the property of discriminative
ranking generation. The proof is shown in the Apendix.

\section{Some recommendations for selecting QFMs for applications}

In table \ref{TableSummary} we synthesize the behavior of the QFMs
$\mathcal{F}^{MD}$, $\mathcal{F}^{I}$, $\mathcal{F}^{A}$, $\mathcal{M}$,
$\mathcal{M}_{CX}$ and $\mathcal{F}_{owa}$ with respect to the set of
additional criteria we have presented. This allow to establish some
recommendations for the selection of convenient models for applications:

\begin{enumerate}
\item In applications that require a fine distinction between objects (e.g.,
ranking applications) only models $\mathcal{F}^{I}$ and $\mathcal{F}^{A}$
should be used for non unary quantifiers. In the unary case $\mathcal{F}^{MD}$
and $\mathcal{F}_{owa}$ coincide with the $\mathcal{F}^{I}$ model for
increasing quantifiers, and are also acceptable.

\item In applications in which aggregative behavior is not acceptable,
$\mathcal{F}^{A}$ should be avoided.

\item For maximal coherence with linguistic criteria, models $\mathcal{M}%
_{CX}$ and $\mathcal{F}^{A}$ are the preferred ones. Models $\mathcal{M}$ and
$\mathcal{F}_{owa}$ show a good behavior as well. Model $\mathcal{F}^{I}$,
being inferior to DFSs with respect to linguistic coherence, conserves
linguistic transformations of the `Aristotelian square' in the finite case.

\item If propagation of fuzziness is required, the only viable options are
$\mathcal{M}$ and $\mathcal{M}_{CX}$.

\item In order to preserve the intuitions underneath of the identity
quantifier, guaranteeing that a linear increase in the inputs produces a
linear increase in the outputs, models $\mathcal{F}^{A}$,$\mathcal{F}^{MD}$,
$\mathcal{F}^{I}$ or $\mathcal{F}_{owa}$ should be selected.

\item When taken into account the behavior of QFMs over quantified partitions,
if we expect more undefined results for fuzzier fuzzy sets, standard DFSs
should be used. In the case of prefering that QFMs could be interpreted as
probabilities over quantified labels, distributing the `degree of fulfilment'
between the different labels, the convenient models are $\mathcal{F}^{MD}$,
$\mathcal{F}^{I}$ and $\mathcal{F}^{A}$.
\end{enumerate}

$\mathcal{\smallskip}$Summing up, the model $\mathcal{F}^{A}$ is a really
convenient model for all the applications in which aggregative behavior is not
an impediment. $\mathcal{M}_{CX}$ is the perfect model for applications in
which preservation of fuzziness properties is required, but presents the
handicap that is very inadequate for ranking applications and it does not
maintain the linguistic intuitions under the `identity quantifier'.
Additionally, it has been proved that the $\mathcal{M}_{CX}$ models presents a
very stable behavior \cite[section 7.12]{Glockner06Libro}, which assures a
certain insensitivity against modifications in the memberships degrees.

If we need a model guaranteeing a fine distinction between objects but
avoiding aggregative behavior, the best option is the $\mathcal{F}^{I}$ model.
$\mathcal{F}^{I}$ also guarantees linguistic intuitions associated to the
identity quantifier, allows to interpret quantified partitions as
probabilities, and for fuzzy sets whose membership degrees are maximally
distributed over the referential set, evaluation results provided by
$\mathcal{F}^{I}$ tend to the area of the quantifier. Moreover, $\mathcal{F}%
^{I}$ preserves internal an external negation properties (this last property
in the finite case), assuring the conservation of the linguistic relations of
the `Aristotelian square'. Although $\mathcal{F}^{I}$ is not a DFS, it is a
remarkable model that presents a great equilibrium between the fulfillment of
the different criteria.

$\mathcal{F}^{MD}$, sharing some of the behavior of the $\mathcal{F}^{I}$
model, is not adequate for fine differentiation of objects in the binary case.
We consider linguistic behavior of $\mathcal{F}^{I}$ model superior to the
linguistic behavior of $\mathcal{F}^{MD}$, as this last model does not
preserve linguistic transformations of the Aristotelian square.

$\mathcal{M}$ model shares most of the behavior of the $\mathcal{M}_{CX}$
model, presenting the same problems but loosing some properties, as Fuzzy
Argument Insertion.

$\mathcal{F}_{owa}$ model has been presented as the paradigmatic example of
standard DFS convenient for ranking applications, but we have seen that this
model is not adequate for achieving a fine differentiation between objects
with binary quantifiers. However, if we were interested in preserving all the
properties of standard DFS guaranteeing some discriminative power, then the
$\mathcal{F}_{owa}$ model is the convenient option.

Finally, the way in which QFMs behave over quantified partitions can guide us
in our decision between standard DFSs and the remaining models. Standard DFS
will tend to produce more undefined results (in the sense of closeness to
$\frac{1}{2}$) for fuzzier fuzzy sets (in the sense of closeness to $\frac
{1}{2}$ of their membership degrees). $\mathcal{F}^{A}$, $\mathcal{F}^{MD}$
and $\mathcal{F}^{I}$ generate results that can be interpreted as
probabilities, dividing the `evaluation weight' between the different
quantifiers in the partition. $\mathcal{F}^{MD}$ and $\mathcal{F}^{I}$ also
preserve the intuition of `weight of the quantifier' (in the sense of the
coverage of the quantification universe by the labels) for a perfect
distribution of membership degrees. That is, $\mathcal{F}^{MD}$ and
$\mathcal{F}^{I}$ tend to produce a result proportional to the area of the
quantifier for fuzzy sets whose membership degrees tend to be equally
distributed over $\left[  0,1\right]  $.%

\begin{table}[tbp] \centering
\begin{tabular}
[c]{|l|l|l|l|l|l|l|}\hline
& $\mathcal{F}^{MD}$ & $\mathcal{F}^{I}$ & $\mathcal{F}^{A}$ & $\mathcal{M}$ &
$\mathcal{M}_{CX}$ & $\mathcal{F}_{owa}$\\\hline
{\small Linguistic Compatibility} & {\small partial} & {\small partial} &
{\small DFS+FAI} & {\small DFS} & {\small DFS+FAI} & {\small DFS}\\\hline
{\small Aggregative behavior} & {\small No} & {\small No} & {\small Yes} &
{\small No} & {\small No} & {\small No}\\\hline
{\small Identity Quantifier} & {\small Yes} & {\small Yes} & {\small Yes} &
{\small No} & {\small No} & {\small Yes}\\\hline
{\small Propagation of Fuzziness} & {\small No} & {\small No} & {\small No} &
{\small Yes} & {\small Yes} & {\small No}\\\hline
{\small Quantified Partitions} & {\small Pr} & {\small Pr} & {\small Pr} &
{\small Ind} & {\small Ind} & {\small Ind}\\\hline
{\small Fine differentiation} & {\small No} & {\small Yes} & {\small Yes} &
{\small No} & {\small No} & {\small No}\\\hline
\end{tabular}
\caption{Summary of the behaviour of the QFMs. FAI: Fuzzy Argument Insertion, Pr: probability interpretation, Ind: tendency to 0.5 evaluation results \label{TableSummary}}%
\end{table}%

\section{Conclusions}

In this work we have advanced in the definition of some criteria to provide a
better understanding of the behavior of the
most significative QFMs. First, we have compared the selected QFMs against the
main set of properties presented in \cite{Glockner06Libro}, with the objective
of clarifying the differences that these models present with respect to the
properties proposed in the QFM framework.

After that, we argued that previous considered properties, while being really
convenient to separate `good quantification models' from `bad ones', are not
sufficient to clearly distinguish between the set of analyzed QFMs, and
specially, to help potential users in the process of selecting the most
convenient model for a specific application.

In order to advance in this problem, we have introduced a new set of criteria,
specially designed to differentiate the behavior of the analyzed models. An
in-depth comparative analysis of the main models has been performed with
respect to this new set of criteria. Based on this analysis we have
established some recommendations to guide in the selection of the more
adequate model for specific practical applications.

As future work, we consider relevant the possibility of defining new oriented
criteria, focused on specific applications.


\section*{Appendix}

\textbf{Discriminative ranking generation, model} $\mathcal{F}^{I}$

\begin{proof}
Intuitively, when we increase the fulfillment of a property $\mu_{X^{o_{i}}%
}\left(  p_{j}\right)  $ associated to a weight greater than $0$ from $a$ to
$b$, we are adding an element to the alpha-cuts in the range $\left(
a,b\right]  $. As the weight of $p_{j}$ is greater than $0$, the relative
cardinality with respect to the alpha-cuts of $W$ containing $p_{j}$ will
increase.\newline In detail, let be $\mu_{W}\left(  p_{j}\right)  =c>0$ and
$\mu_{X^{o_{i}}}\left(  p_{j}\right)  =a$ the fulfillment of the criteria
$p_{j}$ for the object $i$. Let us consider a second fuzzy set $X^{o_{i}%
^{\prime}}$ such that $\mu_{X^{o_{i}}}\left(  p_{z}\right)  =\mu
_{X^{o_{i}\prime}}\left(  p_{z}\right)  $ for every $z\neq j$, and
$\mu_{X^{o_{i}\prime}}\left(  p_{j}\right)  =b>a$.\newline Then,%
\begin{align*}
\mathcal{F}^{I}\left(  Q_{h}\right)  \left(  W,X^{o_{i}}\right)   &  =\int
_{0}^{1}\int_{0}^{1}Q_{h}\left(  W_{\geq\alpha_{1}},X_{\geq\alpha_{2}}^{o_{i}%
}\right)  d\alpha_{1}d\alpha_{2}\\
&  =\int_{0}^{1}\int_{0}^{a}Q_{h}\left(  W_{\geq\alpha_{1}},X_{\geq\alpha_{2}%
}^{o_{i}}\right)  d\alpha_{1}d\alpha_{2}+\int_{0}^{1}\int_{a}^{b}Q_{h}\left(
W_{\geq\alpha_{1}},X_{\geq\alpha_{2}}^{o_{i}}\right)  d\alpha_{1}d\alpha_{2}\\
&  +\int_{0}^{1}\int_{b}^{1}Q_{h}\left(  W_{\geq\alpha_{1}},X_{\geq\alpha_{2}%
}^{o_{i}}\right)  d\alpha_{1}d\alpha_{2}%
\end{align*}
\newline Expressions $\int_{0}^{1}\int_{0}^{a}Q_{h}\left(  W_{\geq\alpha_{1}%
},X_{\geq\alpha_{2}}^{o_{i}}\right)  d\alpha_{1}d\alpha_{2}$ and $\int_{0}%
^{1}\int_{b}^{1}Q_{h}\left(  W_{\geq\alpha_{1}},X_{\geq\alpha_{2}}^{o_{i}%
}\right)  d\alpha_{1}d\alpha_{2}$ are equal for $o_{i}$ and $o_{i}^{^{\prime}%
}$. With respect to $\int_{0}^{1}\int_{a}^{b}Q_{h}\left(  W_{\geq\alpha_{1}%
},X_{\geq\alpha_{2}}^{o_{i}}\right)  d\alpha_{1}d\alpha_{2}$, for
\textit{alpha-cuts} in $\left(  0,c\right]  \times\left(  a,b\right]  $:
\[
Q_{h}\left(  W_{\geq\alpha_{1}},X_{\geq\alpha_{2}}^{o_{i}}\right)
>Q_{h}\left(  W_{\geq\alpha_{1}},X_{\geq\alpha_{2}}^{o_{i}^{\prime}}\right)
\]
as $p_{j}\in W_{\geq\alpha_{1}}$, and $p_{j}\in X_{\geq\alpha_{2}}%
^{o_{i}^{\prime}}$ but $p_{j}\notin X_{\geq\alpha_{2}}^{o_{i}}$. And then
$\mathcal{F}^{I}\left(  Q_{h}\right)  \left(  W,X^{o_{i}}\right)
<\mathcal{F}^{I}\left(  Q_{h}\right)  \left(  W,X^{o_{i}^{\prime}}\right)  $.
\end{proof}

\textbf{Discriminative ranking generation, model} $\mathcal{F}^{A}$

\begin{proof}
Again, let be $\mu_{W}\left(  p_{j}\right)  =c>0$ and $\mu_{X^{o_{i}}}\left(
p_{j}\right)  =a$ the fulfillment of the criteria $p_{j}$ for the object $i$.
Let us consider a second fuzzy set $X^{o_{i}^{\prime}}$ such that
$\mu_{X^{o_{i}}}\left(  p_{z}\right)  =\mu_{X^{o_{i}\prime}}\left(
p_{z}\right)  $ for every $z\neq j$, and $\mu_{X^{o_{i}\prime}}\left(
p_{j}\right)  =b>a$. We are trying to prove that:\newline%
\begin{align*}
\mathcal{F}^{A}\left(  Q_{h}\right)  \left(  W,X^{o_{i}}\right)   &
=\sum_{Y_{1}\in\mathcal{P}\left(  E\right)  }\sum_{Y_{2}\in\mathcal{P}\left(
E\right)  }m_{W}\left(  Y_{1}\right)  m_{X^{o_{i}}}\left(  Y_{1}\right)
Q_{h}\left(  Y_{1},Y_{2}\right) \\
&  <\sum_{Y_{1}\in\mathcal{P}\left(  E\right)  }\sum_{Y_{2}\in\mathcal{P}%
\left(  E\right)  }m_{W}\left(  Y_{1}\right)  m_{X^{o_{i}\prime}}\left(
Y_{1}\right)  Q_{h}\left(  Y_{1},Y_{2}\right)  =\mathcal{F}^{A}\left(
Q_{h}\right)  \left(  W,X^{o_{i}^{\prime}}\right)
\end{align*}
\newline Making some computations with $\mathcal{F}^{A}\left(  Q_{h}\right)
\left(  W,X^{o_{i}}\right)  $ we obtain\footnote{By $E\backslash\left\{
e_{i}\right\}  $ we denote $E\cap\overline{\left\{  e_{i}\right\}  }$; that
is, the set $E$ without the element $e_{i}$. For fuzzy sets, $X\backslash
\left\{  p_{i}\right\}  $ is the projection of $X$ eliminating the $p_{i}$
element. Then, in $m_{X\backslash\left\{  p_{i}\right\}  }\left(  Y\right)  $
the element $p_{i}$ is not taken into account in the computation of the
probability mass of $Y$.}:
\begin{align}
&  \mathcal{F}^{A}\left(  Q_{h}\right)  \left(  W,X^{o_{i}}\right) \nonumber\\
&  =\sum_{Y_{1}\in\mathcal{P}\left(  E\right)  }\sum_{Y_{2}\in\mathcal{P}%
\left(  E\right)  }m_{W}\left(  Y_{1}\right)  m_{X^{o_{i}}}\left(
Y_{2}\right)  Q_{h}\left(  Y_{1},Y_{2}\right) \nonumber\\
&  =\sum_{Y_{1}\in\mathcal{P}\left(  E\backslash\left\{  p_{i}\right\}
\right)  }\sum_{Y_{2}\in\mathcal{P}\left(  E\backslash\left\{  p_{i}\right\}
\right)  }\left(  1-c\right)  m_{W\backslash\left\{  p_{i}\right\}  }\left(
Y_{1}\right)  \left(  1-a\right)  m_{X^{o_{i}}\backslash\left\{
p_{i}\right\}  }\left(  Y_{2}\right)  Q_{h}\left(  Y_{1},Y_{2}\right)
\label{DFA15}\\
&  +\sum_{Y_{1}\in\mathcal{P}\left(  E\backslash\left\{  p_{i}\right\}
\right)  }\sum_{Y_{2}\in\mathcal{P}\left(  E\right)  |p_{i}\in Y_{2}}a\left(
1-c\right)  m_{W\backslash\left\{  p_{i}\right\}  }\left(  Y_{1}\right)
m_{X^{o_{i}}\backslash\left\{  p_{i}\right\}  }\left(  Y_{2}\right)
Q_{h}\left(  Y_{1},Y_{2}\right) \label{DFA2}\\
&  +\sum_{Y_{1}\in\mathcal{P}\left(  E\right)  |p_{i}\in Y_{1}}\sum_{Y_{2}%
\in\mathcal{P}\left(  E\backslash\left\{  p_{i}\right\}  \right)  }c\left(
1-a\right)  m_{W\backslash\left\{  p_{i}\right\}  }\left(  Y_{1}\right)
m_{X^{o_{i}}\backslash\left\{  p_{i}\right\}  }\left(  Y_{2}\right)
Q_{h}\left(  Y_{1},Y_{2}\right) \label{DFA3}\\
&  +\sum_{Y_{1}\in\mathcal{P}\left(  E\right)  |p_{i}\in Y_{1}}\sum_{Y_{2}%
\in\mathcal{P}\left(  E\right)  |p_{i}\in Y_{2}}ca\times m_{W\backslash
\left\{  p_{i}\right\}  }\left(  Y_{1}\right)  m_{X^{o_{i}}\backslash\left\{
p_{i}\right\}  }\left(  Y_{2}\right)  Q_{h}\left(  Y_{1},Y_{2}\right)
\label{DFA4}%
\end{align}
but if $p_{i}\notin Y_{1}$, then the relative cardinality $\frac{\left\vert
Y_{1}\cap C\right\vert }{\left\vert Y_{1}\right\vert }=\frac{\left\vert
Y_{1}\cap\left(  C\cup\left\{  p_{i}\right\}  \right)  \right\vert
}{\left\vert Y_{1}\right\vert }$ for $C\in\mathcal{P}\left(  E\backslash
\left\{  p_{i}\right\}  \right)  $. Then, the sum of expressions \ref{DFA15}
and \ref{DFA2}:%
\begin{align*}
&  \sum_{Y_{1}\in\mathcal{P}\left(  E\backslash\left\{  p_{i}\right\}
\right)  }\sum_{Y_{2}\in\mathcal{P}\left(  E\right)  \backslash\left\{
p_{i}\right\}  }\left(  1-c\right)  m_{W\backslash\left\{  p_{i}\right\}
}\left(  Y_{1}\right)  \left(  1-a\right)  m_{X^{o_{i}}\backslash\left\{
p_{i}\right\}  }\left(  Y_{2}\right)  Q_{h}\left(  Y_{1},Y_{2}\right) \\
&  +\sum_{Y_{1}\in\mathcal{P}\left(  E\backslash\left\{  p_{i}\right\}
\right)  }\sum_{Y_{2}\in\mathcal{P}\left(  E\right)  |p_{i}\in Y_{2}}a\left(
1-c\right)  m_{W\backslash\left\{  p_{i}\right\}  }\left(  Y_{1}\right)
m_{X^{o_{i}}\backslash\left\{  p_{i}\right\}  }\left(  Y_{2}\right)
Q_{h}\left(  Y_{1},Y_{2}\right) \\
&  =\sum_{Y_{1}\in\mathcal{P}\left(  E\backslash\left\{  p_{i}\right\}
\right)  }\sum_{C\in\mathcal{P}\left(  E\backslash\left\{  p_{i}\right\}
\right)  }\left(  1-c\right)  m_{W\backslash\left\{  p_{i}\right\}  }\left(
Y_{1}\right)  m_{X^{o_{i}}\backslash\left\{  p_{i}\right\}  }\left(  C\right)
Q_{h}\left(  Y_{1},C\right)
\end{align*}
is not affected by the modification of $\mu_{X^{o_{i}}}\left(  p_{j}\right)
$; that is, it will coincide with the equivalent expression for $\mathcal{F}%
^{A}\left(  Q_{h}\right)  \left(  W,X^{o_{i}\prime}\right)  $.\newline We will
focus now in \ref{DFA3} and \ref{DFA4}. For $\mathcal{F}^{A}\left(
Q_{h}\right)  \left(  W,X^{o_{i}\prime}\right)  $, equivalent expression of
\ref{DFA3} and \ref{DFA4} are obtained by substituting $\left(  1-a\right)  $
and $a$ by $\left(  1-b\right)  $ and $b$, respectively; that is, we reduced
$\left(  1-a\right)  $ by an $\left(  b-a\right)  $ factor and we increase $a$
by an $\left(  b-a\right)  $ factor. As $h\left(  x\right)  $ is increasing,
$\left(  1-a\right)  h\left(  x\right)  +ah\left(  y\right)  <\left(
1-b\right)  h\left(  x\right)  +bh\left(  y\right)  $ for $b>a$, $y>x$. Thus,
it is trivial to see that \ref{DFA3} and \ref{DFA4} are lesser than the
equivalent expressions for $\mathcal{F}^{A}\left(  Q_{h}\right)  \left(
W,X^{o_{i}\prime}\right)  $.
\end{proof}

\end{document}